\documentclass[10pt, compsoc, journal]{IEEEtran}

\usepackage{amsmath,amsfonts,amssymb}
\usepackage{mathtools,amsthm}
\usepackage{array}
\usepackage[caption=false,font=normalsize,labelfont=sf,textfont=sf]{subfig}
\usepackage{textcomp}
\usepackage{stfloats}
\usepackage{url}
\usepackage{verbatim}
\usepackage{graphicx}
\usepackage{cite}
\usepackage{booktabs,multirow,rotating}
\usepackage{xcolor}
\usepackage{stfloats}
\usepackage{cuted}
\usepackage{enumitem}
\usepackage{multirow}
\usepackage{booktabs}
\usepackage{tabularx}
\usepackage{array}
\usepackage{hyperref}

\theoremstyle{plain}

\newtheorem{proposition}{Proposition}

\theoremstyle{definition}

\newtheorem{assumption}{Assumption}

\theoremstyle{remark}
\newtheorem{remark}{Remark}

\renewcommand{\paragraph}[1]{
    \par\noindent\textbf{#1}\quad
}
\title{How Far Can Untrained Neural Networks Go in Anomaly Detection? From Tables to Pixels}
\author{
Hyojun Yi,
Geunyoung Kim,
Hogeon Seo,
and 
Seunghyoung Ryu, \IEEEmembership{Member, IEEE}
\thanks{Hyojun Yi and Geunyoung Kim contributed equally to this work.}%
\thanks{H. Yi, G. Kim, and S. Ryu are with the Department of Artificial Intelligence and Robotics,
Sejong University, Seoul, Republic of Korea
(e-mail: black99y@sju.ac.kr; kky5572558@sju.ac.kr; shryu@sejong.ac.kr).}%
\thanks{H. Seo is with the Applied AI Section, Korea Atomic Energy Research Institute,
Daejeon, Republic of Korea
(e-mail: hogeony@kaeri.re.kr).}%
\thanks{Corresponding author: Seunghyoung Ryu (e-mail: shryu@sejong.ac.kr).}
\thanks{The source code for PRISM-T and PRISM-V is publicly
available at \url{https://github.com/ideal-lab-sejong/PRISM_T}
and
\url{https://github.com/ideal-lab-sejong/PRISM_V},
respectively.}
\thanks{This work has been submitted to the IEEE for possible publication. Copyright may be transferred without notice, after which this version may no longer be accessible.}
}

\begin{document}
\maketitle

\begin{abstract}
Anomaly detection (AD) is often framed as a representation-learning problem, where strong performance is expected to require learned features or AD-specific optimization.
However, this view can overlook the intrinsic separability of anomalies from normal references, even in the absence of learned representations.
We therefore investigate how far untrained neural networks can go in anomaly detection.
To this end, we propose \textit{PRISM}, a framework combining fixed random nonlinear features with normal-reference Mahalanobis scoring for tabular detection and dense visual localization.
Across 47 tabular datasets in ADBench, \textit{PRISM-T} achieves the highest mean AUROC (86.4\%) and the second-highest mean AUPRC (65.9\%) among the compared methods.
On MVTec AD and VisA, covering 27 visual categories, \textit{PRISM-V} achieves mean image-level and pixel-level AUROCs of 90.4\% and 96.4\%, respectively.
Notably, both models achieve these results without neural weight optimization or pretrained representations.
A theoretical analysis further characterizes how nonlinear transformations affect the mean and covariance contributions to the Mahalanobis score gap, while empirical analyses examine the resulting score separability.
These findings establish \textit{PRISM} as both a competitive anomaly detector and a prior-to-learning baseline for assessing the added value of learned representations in anomaly detection.
\end{abstract}

\begin{IEEEkeywords}
Anomaly detection, random neural features, untrained neural networks, Mahalanobis scoring, visual anomaly localization.
\end{IEEEkeywords}

\section{Introduction}
\label{sec1.intro}

\IEEEPARstart{A}{nomaly} detection (AD) is fundamentally a problem of identifying deviation from normality.
Unlike conventional predictive learning, anomalies need not be observed during model construction; a model may instead characterize normal data and determine how strongly a new observation departs from that reference.
Modern AD methods nevertheless devote substantial modeling effort to constructing representations or scoring spaces in which deviations from normality become easier to expose.
Depending on the formulation, this may involve learning compact representations of normal data through reconstruction or one-class objectives\cite{zhou2017anomaly,ruff2018deep}, characterizing local neighborhoods or normal-data geometry\cite{goodge2022lunar,roth2022patchcore}, modeling the normal distribution through generative or score-based objectives\cite{livernoche2024dte,li2026scalable}, or leveraging pretrained representations in which semantic or structural deviations become more discernible \cite{defard2021padim,damm2025anomalydino,lendering2026subspacead}.
Such approaches reflect a common emphasis on learned representations as a key source of anomaly separability.

Yet a simpler possibility remains.
If abnormality is defined relative to normal data, some anomaly evidence may already be intrinsic to the data, even before representation learning.
Intuitively, we often sense that ``something looks wrong'' before we can precisely identify which rule or regularity has been violated.
We therefore ask a basic but easily overlooked question: how far can anomaly detection go before learning?

This prior-to-learning perspective was partially explored in an earlier study on untrained neural networks (UNNs), which showed that randomly initialized neural features combined with Mahalanobis scoring can detect anomalies without neural weight optimization \cite{ryu2024unn}.
However, the evidence remained limited in both breadth and granularity: the tabular evaluation covered only a small set of datasets, while the visual experiments were restricted to flattened MNIST and F-MNIST with image-level scoring.
It therefore remained unclear whether prior-to-learning separability could support competitive detection across diverse tabular datasets or enable dense visual localization.

To investigate this question, we propose \textsc{PRISM} (Prior-to-learning Random Intrinsic Scoring Model), a framework that exploits intrinsic anomaly evidence through frozen, randomly initialized neural features and normal-reference Mahalanobis scoring.
At its core, \textsc{PRISM} is built upon a simple and lightweight PRISM block consisting of a randomly initialized linear operator, normalization, and kinked nonlinear activation.
This design principle is instantiated in two modality-specific models: \textsc{PRISM-T} for global tabular anomaly detection and \textsc{PRISM-V} for dense visual anomaly localization.
Neither model optimizes neural weights or relies on pretrained representations.
For tabular AD, \textsc{PRISM-T} achieves competitive performance against recent state-of-the-art detectors across 47 ADBench datasets.
For visual AD, \textsc{PRISM-V} achieves strong image-level detection and extends prior-to-learning AD to spatially coherent pixel-level localization on industrial anomaly benchmarks.
These results demonstrate the potential of prior-to-learning anomaly detection across tabular and visual domains, establishing \textsc{PRISM} as a competitive detector built on lightweight, untrained neural feature extractors and a reference for assessing the added value of learned representations.

Our main contributions are summarized as follows.

\begin{itemize}[leftmargin=*, labelindent=0pt, nosep]

\item To examine how far anomaly detection can go before learning, we introduce \textsc{PRISM}, a framework combining random neural features with Mahalanobis scoring.
Through modality-specific implementations, \textsc{PRISM} extends untrained neural anomaly detection from tabular data to visual anomaly detection and localization.

\item We conduct a comprehensive evaluation from tables to pixels, spanning 47 tabular datasets from ADBench and 27 visual categories from MVTec AD and VisA.
For tabular AD, \textsc{PRISM-T} achieves the highest mean AUROC and the second-highest mean AUPRC among the compared detectors, while \textsc{PRISM-V} demonstrates strong image-level detection and pixel-level localization.
Both models achieve these results using fully untrained neural networks with substantially fewer parameters than the compared state-of-the-art neural AD models.

\item We investigate the mechanisms underlying prior-to-learning anomaly separability through theoretical and empirical analyses.
The theory characterizes how nonlinear feature transformations affect the mean and covariance contributions to the expected Mahalanobis score gap.
Experiments further examine the resulting score separability and its relationship with detection performance across both modalities.

\end{itemize}

The remainder of this paper is organized as follows.
Section~\ref{sec2.relatedwork} reviews related work.
Sections~\ref{sec3.ptlanomalydetection} and~\ref{sec4.theory} introduce the PRISM framework and provide a theoretical analysis of anomaly score separation.
Experimental results are presented in Section~\ref{sec5.experiments}, with further analysis in Section~\ref{sec6.analysis}.
Finally, Section~\ref{sec7.conclusion} concludes the paper.

\section{Related Work}
\label{sec2.relatedwork}
\subsection{Anomaly detection and tabular benchmarks}
Tabular AD has long relied on diverse classical approaches, with methods such as Isolation forest, kNN, LOF, and OCSVM continuing to serve as competitive baselines \cite{liu2008iforest,han2022adbench}.
ADBench \cite{han2022adbench} further standardized large-scale tabular evaluation and highlighted the difficulty arising from substantial variation in data structure and anomaly characteristics.
Alongside these classical approaches, neural methods have increasingly been explored to capture more complex nonlinear structures in tabular data.
Recent neural tabular AD methods extend these classical principles by learning representations, nonlinear dependencies, and relational structures.
Some methods learn representations through contrastive, masked, or disentangled objectives \cite{shenkar2022icl,mcm2024,disentad2025}; others refine outlier scoring through neighborhood- or graph-based modeling, as in LUNAR \cite{goodge2022lunar}.
State-of-the-art approaches have increasingly adopted perturbation-based objectives, modeling normality through responses to noise or time-conditioned transformations.
This trend spans diffusion-time estimation, score matching, and flow-matching-inspired approaches, as exemplified by DTE~\cite{livernoche2024dte}, K-DSM~\cite{livernoche2026kurtosis}, and TCCM~\cite{li2026scalable}.
Emerging work further explores LLM- or foundation-model-inspired tabular detectors \cite{tsai2025anollm}.
Despite their methodological diversity, most recent high-performing approaches rely on neural modeling and task-specific training to construct anomaly representations or scoring functions.

\subsection{Visual anomaly detection and localization}
Early deep visual AD methods often treated an image as a single observation, using reconstruction, generative modeling, or one-class representation learning to produce image-level anomaly scores or coarse residual-based maps \cite{schlegl2019fanogan,akcay2018ganomaly,akccay2019skip,ruff2019deep}. 
Industrial visual AD later shifted its emphasis toward dense localization, modeling normality at the patch or spatial-location level rather than treating each image as a single observation. 
Representative methods such as SPADE \cite{cohen2020sub}, PaDiM \cite{defard2021padim}, and PatchCore \cite{roth2022patchcore} established a pretrained feature paradigm based on nearest-neighbor retrieval, location-wise Gaussian modeling, or memory-bank comparison.
This reliance on pretrained visual representations has more recently extended to foundation models, including vision-language models and self-supervised visual backbones, as exemplified by WinCLIP~\cite{jeong2023winclip}, AnomalyDINO~\cite{damm2025anomalydino}, and SubspaceAD~\cite{lendering2026subspacead}.
Although such methods are commonly described as \emph{training-free}, they avoid task-specific feature learning while still relying on representations acquired through large-scale pretraining.
We use the term \emph{prior-to-learning} to distinguish the setting considered in \textsc{PRISM}, where the neural feature extractor is randomly initialized and remains fixed without feature learning or pretraining.
Accordingly, \textsc{PRISM-V} investigates visual anomaly detection and dense localization using only untrained visual representations.

\subsection{Random features and untrained networks}
Although random features may seem to lack learned semantic or task-specific information, fixed random transformations have long been studied for the useful geometric structure they can induce.
For example, the Johnson--Lindenstrauss lemma shows that random linear projections can approximately preserve pairwise Euclidean distances~\cite{johnson1984extensions}.
Likewise, randomly weighted neural networks have been studied as nonlinear embeddings that reshape distances and angles \cite{giryes2016deep}. 
This geometric perspective has also been extended to randomly initialized CNNs, including the effects of convolution and ReLU on angular preservation and contraction \cite{nachum2022johnson}.
These studies establish that random neural mappings can preserve or reshape useful geometric structure, but do not directly address whether the resulting representations expose anomaly separability.
In the tabular domain, UNN~\cite{ryu2024unn} provided empirical evidence that nonlinear features from a randomly initialized multilayer perceptron (MLP), combined with Mahalanobis scoring, can support competitive anomaly detection without training.
DIF~\cite{xu2023deep} employs an ensemble of randomly initialized MLPs to construct nonlinear feature spaces where isolation forests are applied for anomaly scoring.
In visual AD, prior studies have noted that random weights can yield above-chance performance~\cite{chiche2024pre,zhang2025exploring}, but only in auxiliary or ablation analyses rather than as systematic investigations of fully untrained detection.
Despite these findings, whether prior-to-learning representations can support competitive detection across diverse tabular problems and dense visual anomaly localization remains underexplored.
\textsc{PRISM} extends this line of inquiry through a unified investigation of large-scale tabular detection and dense visual localization, together with modality-specific design and a score-level analysis of prior-to-learning separability.

\begin{figure*}[!ht]
\centering
\includegraphics[width=\textwidth]{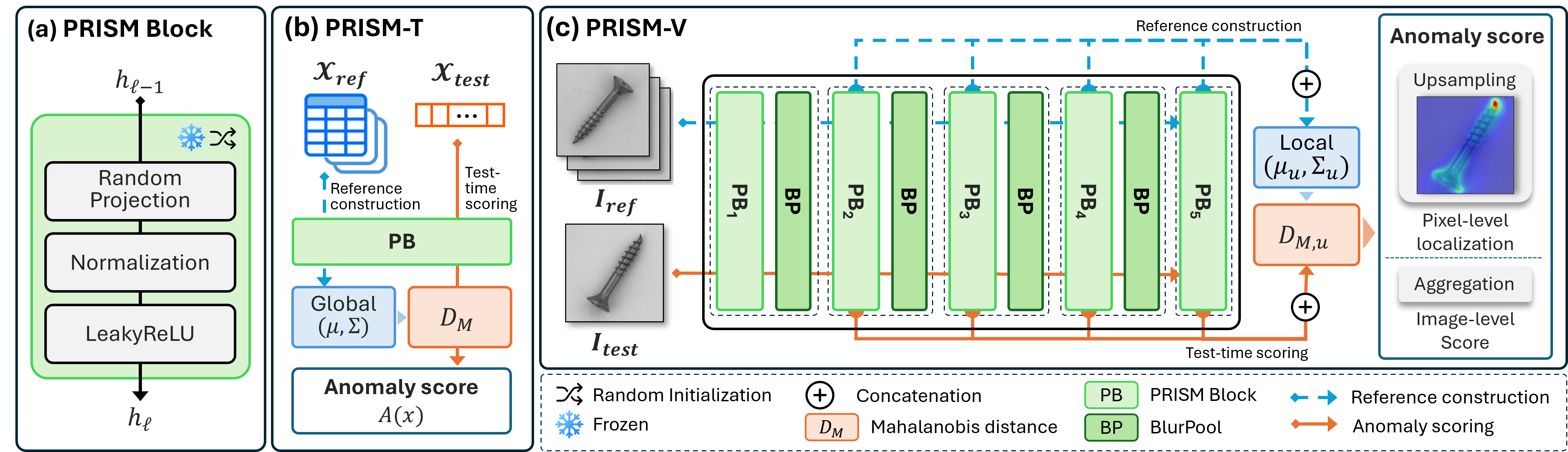}
\caption{PRISM framework.
(a) Each PRISM block (PB) is randomly initialized once and kept fixed.
(b) PRISM-T uses a single PB, and scores global Mahalanobis deviation from the closed-form reference statistics. (c) PRISM-V interleaves five PBs with BlurPool downsampling, aligns and concatenates PBs 2--5, and produces a dense anomaly map.
}
\label{fig:framework}
\end{figure*}

\section{Prior-to-Learning Anomaly Detection}
\label{sec3.ptlanomalydetection}
\subsection{Problem Setting}
We consider semi-supervised AD, where only normal samples are available during model construction.
Let $\mathcal{X}_{\mathrm{ref}}=\{x_i\}_{i=1}^{N}$ denote the normal reference set.
Given an observation $x$, the goal is to assign an anomaly score $A(x)\in\mathbb{R}$, where larger values indicate greater abnormality.

In tabular AD, each observation is a vector $x\in\mathbb{R}^{d}$.
For an image $I$, let $\Omega$ denote a spatial scoring grid and $x_u=\operatorname{vec}(\Pi_u(I))$ the receptive-field patch associated with $u\in\Omega$, where $\Pi_u$ extracts the patch centered at $u$.
A feature extractor $g_\theta$ maps an observation to $z=g_\theta(x)\in\mathbb{R}^{m}$.
For images, the same mapping is shared across locations, yielding $z_u=g_\theta(x_u)$.
Applying the common anomaly score across $\Omega$ produces the location-wise score map
\begin{equation}
A(I)=\{A_u(I)\}_{u\in\Omega},
\qquad
A_u(I)=A(x_u).
\end{equation}
This provides a common formulation for tabular and visual AD.
The score map can be mapped to the input resolution for pixel-level localization, while image-level anomaly scores can be obtained by aggregating spatial scores.

\subsection{Formulation of PRISM}

\textsc{PRISM} uses fixed random transformations to reveal intrinsic normal--anomalous separability without learning feature weights.
This principle leads to two modality-specific instantiations: \textsc{PRISM-T} for tabular AD and \textsc{PRISM-V} for location-wise visual AD.
In each instantiation, one or more \emph{PRISM blocks} (PBs) induce a nonlinear random transformation, after which the resulting features are scored by their Mahalanobis distance from reference statistics.
Figure~\ref{fig:framework} summarizes the shared PB design and the modality-specific pipelines of \textsc{PRISM-T} and \textsc{PRISM-V}.

For an input representation $h_{\ell-1}$, the $\ell$-th PB is defined as
a composition of a randomly initialized linear operator, normalization,
and a kinked nonlinearity:
\begin{equation}
\mathrm{PB}_{\ell}(h_{\ell-1})
=
\sigma\!\left(
\mathrm{Norm}
\left(
\mathrm{Lin}_{\theta_\ell}(h_{\ell-1})
\right)
\right),
\label{eq:block}
\end{equation}
where $\mathrm{Lin}_{\theta_\ell}$ denotes a linear operator with weights
$\theta_\ell$, $\mathrm{Norm}$ denotes a normalization layer without learnable affine parameters, and $\sigma$ is the LeakyReLU activation function.
All weights $\theta_\ell$ are sampled once at initialization and kept fixed; no backpropagation, pretraining, or task-specific optimization is applied.

The feature extractor $g_\theta$ is formed from one or more PBs.
For a given scoring unit, the normal-reference features $\mathcal{Z}_{\mathrm{ref}}
=\{g_\theta(x_i)\mid x_i\in\mathcal{X}_{\mathrm{ref}}\}$ define a mean $\mu$ and covariance $\Sigma$.
The Mahalanobis anomaly score is
\begin{equation}
A(x)
=
\sqrt{
\left(g_\theta(x)-\mu\right)^\top
(\Sigma+\lambda I)^{-1}
\left(g_\theta(x)-\mu\right)
},
\label{eq:mahalanobis}
\end{equation}
where $\lambda$ regularizes covariance inversion.
The reference statistics are global for \textsc{PRISM-T} and location-specific for \textsc{PRISM-V}. 
Thus, \textsc{PRISM} assigns a larger anomaly score when the randomly transformed representation deviates further from the normal-reference distribution.

\paragraph{PRISM for Tabular AD:}The tabular version, denoted \textsc{PRISM-T}, applies a single-PB extractor to vector-valued samples and computes one Mahalanobis anomaly score for each instance. 
In this setting, $\mathrm{Lin}(\cdot)$ in Eq.~\eqref{eq:block} is instantiated as a fully connected layer.
Unlike bounded image pixel values, tabular variables may contain meaningful magnitude information.
Instance Normalization (IN)~\cite{ulyanov2016instance} removes sample-specific mean and scale, emphasizing relative feature patterns but potentially suppressing anomalies expressed through unusual magnitudes.
Accordingly, $\mathrm{Norm}(\cdot)$ applies IN to one half of the feature dimensions and Batch Normalization (BN)~\cite{ioffe2015batch} to the other half, preserving both pattern- and magnitude-level deviations. 
BN statistics are estimated once from the normal reference set and fixed during inference.

\paragraph{PRISM for Visual AD:}\textsc{PRISM-V} extends the same PB-based random feature extraction to visual anomaly localization. 
For \textsc{PRISM-V}, $\mathrm{Lin}(\cdot)$ in Eq.~\eqref{eq:block} is instantiated as a convolutional layer, preserving spatial correspondence while sharing the same fixed random filters across locations.
$\mathrm{Norm}(\cdot)$ uses IN.
To capture visual anomalies at different spatial extents, \textsc{PRISM-V} interleaves PBs with anti-aliased BlurPool~\cite{zhang2019making} downsampling to construct multi-scale random descriptors.
Letting $H_0=I$, the resulting hierarchy is defined as
\begin{equation}
F_\ell=\mathrm{PB}_\ell(H_{\ell-1}),
\qquad
H_\ell=\mathrm{BlurPool}(F_\ell).
\label{eq:visual-hierarchy}
\end{equation}
Selected PB outputs are bilinearly aligned to the common scoring grid $\Omega$ and concatenated into the location-wise descriptor $z_u=[\tilde{F}_{\ell,u}]_{\ell\in\mathcal{T}}$, where $\tilde{F}_{\ell}$ is the aligned feature map from stage $\ell$.
At each $u\in\Omega$, \textsc{PRISM-V} estimates $(\mu_u,\Sigma_u)$ and applies Eq.~\eqref{eq:mahalanobis} to $z_u=g_\theta(x_u)$, yielding the location-wise score $A_u(I)=A(x_u)$.
The resulting score map $A(I)=\{A_u(I)\}_{u\in\Omega}$ is upsampled to the input resolution ($224\times224$) and Gaussian-smoothed to produce the final pixel-level anomaly map $A_{\mathrm{pix}}(I)$.
The image-level anomaly score is obtained by applying empirical tail value-at-risk (TVaR) aggregation to $A_{\mathrm{pix}}(I)$.

\section{Analysis of Anomaly Score Separation}
\label{sec4.theory}

\subsection{Score-Level Separation}
Let $X_{\mathrm N}\sim P_{\mathrm N}$ and $X_{\mathrm A}\sim P_{\mathrm A}$ denote normal and abnormal samples, with $c\in\{\mathrm N,\mathrm A\}$ indexing the class. 
For a fixed PB extractor $g$ and observation space $o$, define $X_c^o=X_c$ and $X_c^g=g(X_c)$. 
With space index $r\in\{o,g\}$, class-conditional mean and covariance in space $r$ are $\mu_{c,r}$ and $\Sigma_{c,r}$, respectively.
The normal-class index $\mathrm{N}$ is omitted when no ambiguity arises, so that $\mu_r=\mu_{\mathrm N,r}$ and $\Sigma_r=\Sigma_{\mathrm N,r}$.
For the population analysis, assume that $\Sigma_r$ is nonsingular and set $M_r=\Sigma_r^{-1}$. 
In the implementation, the mean and covariance are estimated from the reference set, and the ridge-regularized metric in Eq.~\eqref{eq:mahalanobis} is retained for numerical stability.

The class-conditional squared Mahalanobis score is
\begin{equation}
S_c^r
=
(X_c^r-\mu_r)^\top M_r(X_c^r-\mu_r),
\qquad c\in\{\mathrm N,\mathrm A\}.
\label{eq:theory-score}
\end{equation}
To quantify normal--abnormal separation in space $r$, define the standardized score separation as
\begin{equation}
G_r
=
\frac{\Delta_r}{\sqrt{V_r}}
=
\frac{\mathbb E[S_{\mathrm A}^r]-\mathbb E[S_{\mathrm N}^r]}
{\sqrt{\operatorname{Var}(S_{\mathrm A}^r)
+\operatorname{Var}(S_{\mathrm N}^r)}},
\qquad V_r>0.
\label{eq:signed-gap}
\end{equation}
Here, $\Delta_r$ is the expected score difference between abnormal and normal samples, and $V_r$ is the sum of their within-class score variances.
Thus, $G_r$ measures the score gap relative to the combined score dispersion, $\sqrt{V_r}$.
Its squared value corresponds to the Fisher discriminant ratio, while the sign of $\Delta_r$ is retained since larger scores conventionally indicate greater abnormality. 
Note that $G_r$ characterizes score separation through distributional moments rather than the pairwise rankings underlying AUROC.

\begin{proposition}[Decomposition of the expected score gap]
\label{prop:score-gap-decomposition}
For the squared Mahalanobis score in Eq.~\eqref{eq:theory-score},
\begin{equation}
\begin{aligned}
\Delta_r
&=
\underbrace{
\operatorname{tr}
\left[
M_r
(\Sigma_{\mathrm{A},r}-\Sigma_r)
\right]
}_{\text{covariance contribution,}\ \Delta_r^{cov}}
\\
&\quad+
\underbrace{
(\mu_{\mathrm{A},r}-\mu_r)^\top
M_r
(\mu_{\mathrm{A},r}-\mu_r)
}_{\text{mean contribution,}\ \Delta_r^{mean}}.
\end{aligned}
\label{eq:score-gap-decomposition}
\end{equation}
\end{proposition}
The result follows from the quadratic-form identity
\begin{equation}
\mathbb E[S_c^r]
=
\operatorname{tr}(M_r\Sigma_{c,r})
+(\mu_{c,r}-\mu_r)^\top M_r(\mu_{c,r}-\mu_r).
\end{equation}
The covariance contribution measures the difference between abnormal and normal covariance weighted by $M_r$. 
The mean contribution is the squared Mahalanobis distance between their means and is therefore nonnegative.
Dividing Eq.~\eqref{eq:score-gap-decomposition}
by $\sqrt{V_r}$ gives an exact population-level decomposition,
\begin{equation}
G_r
= G_r^{\mathrm{cov}} + G_r^{\mathrm{mean}}
= \frac{\Delta_r^{\mathrm{cov}}}{\sqrt{V_r}}
+ \frac{\Delta_r^{\mathrm{mean}}}{\sqrt{V_r}}.
\label{eq:standardized-contributions}
\end{equation}
These terms quantify covariance- and mean-based contributions relative to the combined score dispersion, $\sqrt{V_r}$. 
In particular, changes in $G_r^{\mathrm{cov}}$ need not correspond to changes of the same magnitude or sign in the unstandardized covariance gap.

\begin{remark}[Invariance under invertible linear transformation]
\label{rem:linear-invariance}
Under population-level Mahalanobis scoring, any fixed invertible linear transformation, including a randomly drawn one, preserves the sample-wise scores and hence $G_r$.
This invariance does not generally extend to non-invertible projections, nonlinear transformations, or ridge-regularized finite-sample metrics.
\end{remark}

\subsection{Local Relation-Violating Anomalies}

\begin{assumption}[Local relation-violating anomaly model]
\label{ass:local-anomaly}
A normal sample is represented as
\begin{equation}
X_{\mathrm N}=f(\zeta)+\varepsilon,
\end{equation}
where $f(\zeta)$ represents an underlying nominal relation and $\varepsilon$ captures admissible normal variation. 
An abnormal counterpart is obtained through
\begin{equation}
X_{\mathrm A}=X_{\mathrm N}+\rho,
\qquad \mathbb E[\rho\mid X_{\mathrm N}]=0,
\label{eq:local-anomaly}
\end{equation}
where $\rho$ is a relation-violating residual with conditional
covariance $C_\rho(x)=\operatorname{Cov}(\rho\mid X_{\mathrm N}=x)$.
Let $\bar C_\rho=\mathbb E[C_\rho(X_{\mathrm N})]$.
\end{assumption}

This stylized model isolates anomalies that introduce additional
conditional variation without a systematic conditional mean shift;
it is not intended to characterize every anomaly in the benchmarks.
In the observation space, the zero-conditional-mean property yields
\begin{equation}
\mu_{\mathrm A,o}=\mu_o,
\qquad
\Sigma_{\mathrm A,o}-\Sigma_o=\bar C_\rho\succeq0.
\end{equation}
Consequently, Proposition~\ref{prop:score-gap-decomposition} implies
\begin{equation}
\Delta_o^{\mathrm{mean}}=0,
\qquad
\Delta_o=\Delta_o^{\mathrm{cov}}
=\operatorname{tr}(M_o\bar C_\rho)\geq0,
\label{eq:observation-residual-gap}
\end{equation}
with a strict inequality when $\bar C_\rho\ne0$.
Hence, the raw Mahalanobis score already contains covariance-based anomaly evidence under this model. 
The question is how a fixed nonlinear transformation changes the same evidence and whether that change improves standardized separation.

\subsection{Nonlinear Mechanisms of Score-Gap Variation}
\label{sec:nonlinear-amplification}

For the coupled normal--abnormal samples in
Assumption~\ref{ass:local-anomaly}, define their transformed displacement as
\begin{equation}
D_g=X_{\mathrm A}^g-X_{\mathrm N}^g
=g(X_{\mathrm N}+\rho)-g(X_{\mathrm N}),
\label{eq:nonlinear-displacement}
\end{equation}
so that $X_{\mathrm A}^g=X_{\mathrm N}^g+D_g$.

\begin{proposition}[Covariance contribution under nonlinear transformation]
\label{prop:covariance-mechanism}
Under the above displacement representation,

\begin{equation}
\begin{aligned}
\Sigma_{\mathrm A,g}-\Sigma_g
={}&\operatorname{Cov}(D_g)
+\operatorname{Cov}(X_{\mathrm N}^g,D_g)\\
&+\operatorname{Cov}(D_g,X_{\mathrm N}^g).
\end{aligned}
\label{eq:nonlinear-covariance-decomposition}
\end{equation}
The covariance contribution to the expected score gap is therefore
\begin{align}
\Delta_g^{\mathrm{cov}}
&=
\underbrace{
\operatorname{tr}
\left[
M_g\operatorname{Cov}(D_g)
\right]
}_{\text{displacement spread}}
+
\underbrace{
2\operatorname{tr}
\left[
M_g\operatorname{Cov}(X_{\mathrm N}^g,D_g)
\right]
}_{\text{interaction}} .
\label{eq:nonlinear-covariance-gap}
\end{align}
\end{proposition}

The displacement-spread term is nonnegative and measures additional variation in the Mahalanobis geometry. 
The interaction term captures the dependence between the transformed normal representation and its anomaly-induced displacement and can have either sign. 
Under Assumption~\ref{ass:local-anomaly}, the corresponding observation-space interaction vanishes because $\operatorname{Cov}(X_{\mathrm N},\rho)=0$, but the interaction need not vanish after nonlinear transformation. 
Specifically,
\begin{equation}
\operatorname{tr}\!\left[
M_g\operatorname{Cov}(X_{\mathrm N}^g,D_g)\right]
=\mathbb E\!\left[
(X_{\mathrm N}^g-\mu_g)^\top M_gD_g\right].
\label{eq:interaction-interpretation}
\end{equation}
A positive interaction indicates that transformed displacements tend to align with existing deviations from the normal mean in the Mahalanobis geometry. 

The transformed displacement can also create a feature-mean difference, $\mu_{\mathrm A,g}-\mu_g=\mathbb E[D_g]$, giving the mean contribution
\begin{equation}
\Delta_g^{\mathrm{mean}}
=\mathbb E[D_g]^\top M_g\mathbb E[D_g]\geq0.
\label{eq:nonlinear-mean-contribution}
\end{equation}

The following identity illustrates how an activation used in \textsc{PRISM} can produce a feature-mean difference even when pre-activation means are unchanged.

\begin{proposition}[LeakyReLU-induced feature-mean gap]
\label{prop:lrelu-mean-gap}
Let $U_{\mathrm N}$ and $U_{\mathrm A}$ denote normal and abnormal inputs to LeakyReLU activation $\sigma_\alpha$, where $0\leq\alpha<1$. 
Then
\begin{equation}
\begin{aligned}
&\mathbb E[\sigma_\alpha(U_{\mathrm A})]
-\mathbb E[\sigma_\alpha(U_{\mathrm N})]\\
&\quad=\frac{1+\alpha}{2}
\bigl(\mathbb E[U_{\mathrm A}]-\mathbb E[U_{\mathrm N}]\bigr)\\
&\qquad+\frac{1-\alpha}{2}
\bigl(\mathbb E|U_{\mathrm A}|-\mathbb E|U_{\mathrm N}|\bigr).
\end{aligned}
\label{eq:lrelu-mean-gap}
\end{equation}
\end{proposition}
This follows from
$\sigma_\alpha(u)=\tfrac{1+\alpha}{2}u+
\tfrac{1-\alpha}{2}|u|$.
For example, let $U_{\mathrm N}\sim\mathcal N(0,\sigma^2)$ and
$U_{\mathrm A}=U_{\mathrm N}+\eta$ with independent
$\eta\sim\mathcal N(0,\tau^2)$. Their preactivation means coincide,
but
\begin{equation}
\mathbb E|U_{\mathrm A}|-\mathbb E|U_{\mathrm N}|
=\sqrt{\frac{2}{\pi}}
\left(\sqrt{\sigma^2+\tau^2}-\sigma\right)>0,
\qquad \tau>0.
\label{eq:lrelu-gaussian-example}
\end{equation}
Thus, LeakyReLU can convert a change in feature variation into a nonzero feature-mean difference and hence a nonnegative quadratic mean contribution when retained in the scored representation.
Note that these mechanisms do not require weight optimization, allowing randomly initialized neural networks to modify the expected score gap.

\subsection{Implications for Standardized Score Separation}
The preceding results characterize how nonlinear transformations modify the covariance and mean contributions to the expected score gap. 
However, as shown in Eq.~\eqref{eq:standardized-contributions}, the standardized separation $G_r$ also depends on the combined score dispersion, $\sqrt{V_r}$.
Since the behavior of $V_r$ under nonlinear transformations depends on the underlying distributions, establishing a general improvement in $G_r$ remains challenging.
The resulting separation is therefore examined empirically in the subsequent analysis section.
\begin{table}[!t]
\caption{
Architecture and configuration of
\textsc{PRISM-T} and \textsc{PRISM-V}.
}
\label{tab:prism-config}
\centering
{
\begin{tabular}{lcc}
\toprule
\textbf{Setting} & \textsc{PRISM-T} & \textsc{PRISM-V} \\ 
\midrule
Linear operator
& Fully connected
& Convolution \\

Number of PBs
& 1
& 5 \\

Feature dimension
& 512
& 512 \\
Kernel size & - & [3,3,3,1,1]\\
Initialization & Uniform & He normal\\
Normalization
& Instance + Batch
& Instance \\
Activation
& \multicolumn{2}{c}{LeakyReLU ($\alpha=0.01$)} \\

Mahalanobis $\lambda$
& $10^{-3}$
& $10^{-2}$ \\

Scoring scope
& Global
& Location-wise \\

\bottomrule
\end{tabular}}

\end{table}
\section{Experiments}
\label{sec5.experiments}
\subsection{Experimental Setup}

\paragraph{Datasets and Evaluation:}
We evaluate \textsc{PRISM} on three widely adopted anomaly detection benchmarks: ADBench~\cite{han2022adbench} for tabular AD, and MVTec AD~\cite{bergmann2019mvtec} and VisA~\cite{zou2022visa} for visual AD. 
Tabular experiments cover the 47 classical datasets in ADBench, excluding the 10 additional CV/NLP datasets represented by pretrained embeddings to maintain a prior-to-learning evaluation setting. 
Following recent normal-only protocols, 50\% of the normal samples are randomly selected as the training set for learning-based baselines~\cite{disentad2025,li2026scalable} and as the reference set for \textsc{PRISM-T}. 
The remaining normal samples and all anomalous samples constitute the test set.
AUROC and AUPRC are evaluated, with mean scores and average ranks reported across datasets.

Visual experiments cover 15 categories of industrial objects and textures in MVTec AD and 12 categories in VisA, including complex multi-instance scenes. 
For MVTec AD, the defect-free training split serves as the normal reference set, and evaluation uses the official test split. 
For VisA, we follow the official one-class protocol, using 90\% of normal images for reference and the remaining normals plus all anomalies for testing. Image-level performance is evaluated using AUROC and AUPRC, and pixel-level localization using AUROC and area under the per-region overlap curve (AUPRO) \cite{bergmann2021mvtec}. 
All metrics range from 0 to 1, with higher values indicating better performance. For both tabular and visual evaluations, experimental results are averaged over five random seeds. 
Throughout the tables, the best and second-best results are highlighted in bold and underlined. 
For brevity, some tables use AUC, AUP, and PRO for AUROC, AUPRC, and AUPRO, respectively, with I- and P- indicating image- and pixel-level metrics.

\begin{table}[!t]
\centering
\caption{Summary of the compared anomaly detection methods,
neural network requirements, and scoring mechanisms.}
\label{tab:baseline}

\begingroup
\scriptsize
\setlength{\tabcolsep}{2pt}
\renewcommand{\arraystretch}{0.95}
\renewcommand{\tabularxcolumn}[1]{m{#1}}

\begin{tabular}{p{0.15\columnwidth}p{0.62\columnwidth}ccc}
\toprule[0.9pt]
\textbf{Method} &
\textbf{Description} &
\textbf{NN} &
\textbf{Tr.} &
\textbf{Sc.} \\
\midrule[0.8pt]
\multicolumn{5}{c}{\textbf{Tabular baselines}} \\
\midrule[0.8pt]

KDE &
Kernel density estimation of normal data. &
-- & -- & LIK \\
\midrule[0.2pt]

AE &
Autoencoder trained on normal data. &
$\checkmark$ & T & REC \\
\midrule[0.2pt]

OCSVM &
Maximum-margin one-class decision boundary. &
-- & -- & DIST \\
\midrule[0.2pt]

IForest &
Random isolation trees with path-length scoring. &
-- & -- & ISO \\
\midrule[0.2pt]

DSVDD &
Distance to a learned hypersphere center. &
$\checkmark$ & T & DIST \\
\midrule[0.2pt]

LUNAR &
GNN-weighted nearest-neighbor distances. &
$\checkmark$ & T & NN \\
\midrule[0.2pt]

DIF &
Isolation forests on random neural features. &
$\checkmark$ & -- & ISO \\
\midrule[0.2pt]

UNN &
Mahalanobis distance on random neural features. &
$\checkmark$ & -- & MD \\
\midrule[0.2pt]

DTE-NP &
Diffusion-time estimation via kNN. &
-- & -- & NN \\
\midrule[0.2pt]

Disent-AD &
Disentangled correlations among attribute subsets. &
$\checkmark$ & T & REC \\
\midrule[0.2pt]

AnoDDAE &
Diffusion-scheduled denoising reconstruction. &
$\checkmark$ & T & REC \\
\midrule[0.2pt]

TCCM &
Time-conditioned contraction matching deviation. &
$\checkmark$ & T & MS \\
\midrule[0.2pt]

K-DSM &
Kurtosis-guided score matching and gradient-norm scoring.&
$\checkmark$ & T & MS \\
\midrule[0.2pt]

DenoiserAD &
Perturbation-based denoising deviation. &
$\checkmark$ & T & MS \\

\midrule[0.8pt]
\multicolumn{5}{c}{\textbf{Visual baselines}} \\
\midrule[0.8pt]

VAE &
Variational autoencoder reconstruction. &
$\checkmark$ & T & REC \\
\midrule[0.2pt]

GANomaly &
Encoder--decoder--encoder GAN with latent reconstruction discrepancy. &
$\checkmark$ & T & REC \\
\midrule[0.2pt]

SPADE &
kNN correspondence over multi-scale pretrained
ResNet feature pyramids. &
$\checkmark$ & P & NN \\
\midrule[0.2pt]

PaDiM &
Patch-wise Gaussian modeling of pretrained
ResNet features with Mahalanobis scoring. &
$\checkmark$ & P & MD \\
\midrule[0.2pt]

PatchCore &
Coreset-subsampled memory bank of pretrained
ResNet features with nearest-neighbor scoring. &
$\checkmark$ & P & NN \\
\midrule[0.2pt]

WinCLIP+ &
CLIP-based window-wise text--image alignment
augmented with normal reference features. &
$\checkmark$ & P & MS \\
\midrule[0.2pt]

Anomaly\newline DINO &
Patch-level nearest-neighbor matching on frozen DINOv2 features. &
$\checkmark$ & P & NN \\
\midrule[0.2pt]

Subspace\newline AD &
PCA subspace fitted to frozen DINOv2 patch features,
with reconstruction residual scoring. &
$\checkmark$ & P & REC \\
\bottomrule
\end{tabular}

\par\smallskip
\parbox{\columnwidth}{\scriptsize
NN: neural network usage ($\checkmark$: yes, --: no);
Tr.: neural weight training (T: task-trained,
P: pretrained, --: none);
Sc.: anomaly scoring function
(NN: nearest-neighbor, REC: reconstruction,
DIST: distance or margin, MD: Mahalanobis distance,
ISO: isolation, LIK: likelihood-based,
MS: method-specific).
}
\endgroup
\end{table}
\paragraph{Architecture and Hyperparameters:}
Table~\ref{tab:prism-config} summarizes the modality-specific architectural designs and the resulting hyperparameter settings.
A single global configuration is selected for each modality using three preselected development datasets or categories (\textsc{PRISM-T}: \textit{Pima, Fault}, and \textit{SpamBase}; \textsc{PRISM-V}: \textit{cable, hazelnut}, and \textit{carpet} from MVTec AD) and then fixed across all evaluations without per-dataset tuning.

\paragraph{Baseline Models:} 
Tabular AD performance is compared against 14 detectors, including recent state-of-the-art methods such as DTE-NP~\cite{livernoche2024dte}, AnoDDAE~\cite{sattarov2025diffusion}, TCCM~\cite{li2026scalable}, and DenoiserAD~\cite{hirthdenoising}.
The comparison also covers strong baselines from prior large-scale benchmarks, such as LUNAR~\cite{goodge2022lunar} and KDE~\cite{parzen1962estimation}, as well as methods using untrained neural features, including DIF~\cite{xu2023deep} and UNN~\cite{ryu2024unn}.
Due to computational resource constraints, KDE, OCSVM, and DenoiserAD are fitted using 10,000 randomly selected normal training samples on the nine datasets with over 50,000 samples (\textit{backdoor, smtp, skin, celeba, census, cover, fraud, http, and donors}).
The test sets remain unchanged.

In the visual AD setting, evaluation is conducted against eight baselines, including generative models (VAE~\cite{kingma2014iclr-auto} and GANomaly~\cite{akcay2018ganomaly}) and pretrained-feature methods. 
The latter include conventional CNN-based approaches (SPADE~\cite{cohen2020sub}, PaDiM~\cite{defard2021padim}, and PatchCore~\cite{roth2022patchcore}) and recent foundation-model-based methods (WinCLIP+~\cite{jeong2023winclip}, AnomalyDINO~\cite{damm2025anomalydino}, and SubspaceAD~\cite{lendering2026subspacead}).
For VAE, we adopt a CNN encoder and a symmetric decoder based on the \textsc{PRISM-V} architecture and generate pixel-level anomaly maps from reconstruction errors following prior studies~\cite{baur2021autoencoders,miranda2022detection}.
The hyperparameters of VAE and GANomaly are selected using the same three MVTec AD development categories used for \textsc{PRISM-V} and fixed across both visual benchmarks.
All other baselines use publicly available implementations from PyOD, scikit-learn, or the original studies with their provided or default hyperparameter settings.
Table~\ref{tab:baseline} provides an overview of the baseline methods, their neural network training and anomaly scoring mechanisms.

\paragraph{Implementation Details:} 
For \textsc{PRISM-T}, input features are z-score standardized and passed through a single PB consisting of a frozen fully connected layer, mixed instance--batch normalization, and LeakyReLU. 
All dataset-level statistics are estimated solely from the reference set.
The random weights are initialized from $\mathcal{U}(-1/\sqrt{d},1/\sqrt{d})$.
For \textsc{PRISM-V}, input images are resized to 256 and center-cropped to $224\times224$, a resolution commonly used in visual AD~\cite{defard2021padim,roth2022patchcore}.
Baseline methods retain their original preprocessing and input resolution settings.
The feature extractor comprises five convolutional PBs with BlurPool downsampling, and the convolutional weights follow He normal initialization~\cite{he2015delving}.
Feature taps are fixed at $\mathcal{T}=\{2,3,4,5\}$.
The tapped feature maps are bilinearly upsampled to a $112\times112$ grid and concatenated for location-wise Mahalanobis scoring. 
The reference mean and covariance are estimated per grid location, with covariance matrices averaged over a $3\times3$ window before regularization and inversion.
Gaussian smoothing uses $\sigma=4$, and TVaR aggregation uses a tail ratio of $\rho=1\%$.

\begin{table}[!t]
\centering
\caption{
Tabular AD results on ADBench datasets in AUROC and AUPRC (\%). 
}
\label{tab:tabular}
{
\setlength{\tabcolsep}{2pt}
\begin{tabular}{lcccc cccc}
\toprule
& \multicolumn{4}{c}{AUROC}
& \multicolumn{4}{c}{AUPRC} \\
\cmidrule(lr){2-5}\cmidrule(lr){6-9}
Method & Mean & Std & $p_{\mathrm H}$ & Top-2
       & Mean & Std & $p_{\mathrm H}$ & Top-2 \\
\midrule

\textbf{PRISM-T}
& \textbf{86.4} & $\pm$0.9 & -- & \underline{17}
& \underline{65.9} & $\pm$2.5 & -- & \underline{12} \\

\midrule

KDE$^\dagger$
& 84.2 & $\pm$0.7 & .315 & 10
& 62.3 & $\pm$2.5 & 1 & 9 \\

AE
& 82.9 & $\pm$1.2 & $<$.001 & 6
& 59.4 & $\pm$3.5 & .004 & 4 \\

OCSVM$^\dagger$
& 79.6 & $\pm$0.7 & .002 & 6
& 55.0 & $\pm$2.2 & .003 & 3 \\

IForest
& 79.8 & $\pm$1.6 & $<$.001 & 5
& 49.3 & $\pm$3.4 & $<$.001 & 4 \\

DSVDD
& 74.8 & $\pm$4.8 & $<$.001 & 0
& 48.4 & $\pm$7.0 & $<$.001 & 2 \\

\midrule

LUNAR
& 85.9 & $\pm$0.9 & 1 & 16
& \textbf{68.0} & $\pm$2.7 & 1 & \textbf{18} \\

DIF
& 76.2 & $\pm$1.7 & $<$.001 & 2
& 48.2 & $\pm$2.5 & $<$.001 & 3 \\

UNN
& 83.5 & $\pm$1.1 & .015 & 8
& 61.0 & $\pm$2.4 & .012 & 8 \\

DTE-NP
& \underline{86.3} & $\pm$0.7 & 1 & \textbf{19}
& 65.5 & $\pm$2.2 & 1 & \underline{12} \\

\midrule

Disent-AD
& 76.1 & $\pm$2.5 & $<$.001 & 2
& 51.3 & $\pm$5.2 & $<$.001 & 3 \\

AnoDDAE
& 85.9 & $\pm$1.1 & .315 & 10
& 65.6 & $\pm$3.0 & 1 & 5 \\

TCCM
& 84.7 & $\pm$2.4 & .262 & 12
& 64.5 & $\pm$4.9 & 1 & 10 \\

K-DSM
& 84.6 & $\pm$1.1 & .723 & 13
& 65.2 & $\pm$3.3 & 1 & 9 \\

DenoiserAD$^\dagger$
& \underline{86.3} & $\pm$1.2 & .774 & 10
& 64.5 & $\pm$4.3 & 1 & 11 \\

\bottomrule
\end{tabular}
\par\smallskip
\parbox{\columnwidth}{\scriptsize
Mean and Std denote the averages of performance and per-dataset standard deviations across seeds, respectively.
$p_{\mathrm H}$ is the Holm-adjusted two-sided Wilcoxon signed-rank $p$-value against PRISM-T.
Top-2 counts top-two rankings across datasets. $\dagger$ indicates a 10K training-sample limit on large datasets.
}}
\end{table}

\subsection{Tabular Anomaly Detection}
Table~\ref{tab:tabular} summarizes the results on 47 ADBench datasets in terms of mean performance with standard deviations, statistical significance, and top-two counts.
Despite using entirely untrained neural features, \textsc{PRISM-T} achieves the highest mean AUROC of $86.4\%$ and the second-highest mean AUPRC of $65.9\%$.
It also records the second-highest Top-2 counts, highlighting its competitive performance across both metrics.
Among the top-performing baselines, DTE-NP and DenoiserAD achieve similar mean AUROC scores of $86.3\%$, but lower mean AUPRC, while LUNAR achieves the highest mean AUPRC of $68.0\%$.
Compared with previous untrained neural feature-based methods, \textsc{PRISM-T} improves upon UNN by $3.5\%$ and $8.0\%$, and DIF by $13.4\%$ and $36.7\%$ in AUROC and AUPRC, respectively.
To assess statistical significance, Holm-adjusted Wilcoxon signed-rank test $p$-values
($p_{\mathrm H}$) are reported for comparisons between \textsc{PRISM-T} and each baseline.
The results show no significant differences ($p_{\mathrm H}\geq0.05$) from leading baselines, including DTE-NP, LUNAR, AnoDDAE, TCCM, K-DSM, and DenoiserAD, on either metric.

\begin{figure}[!t]
\centering
\includegraphics[width=\columnwidth]{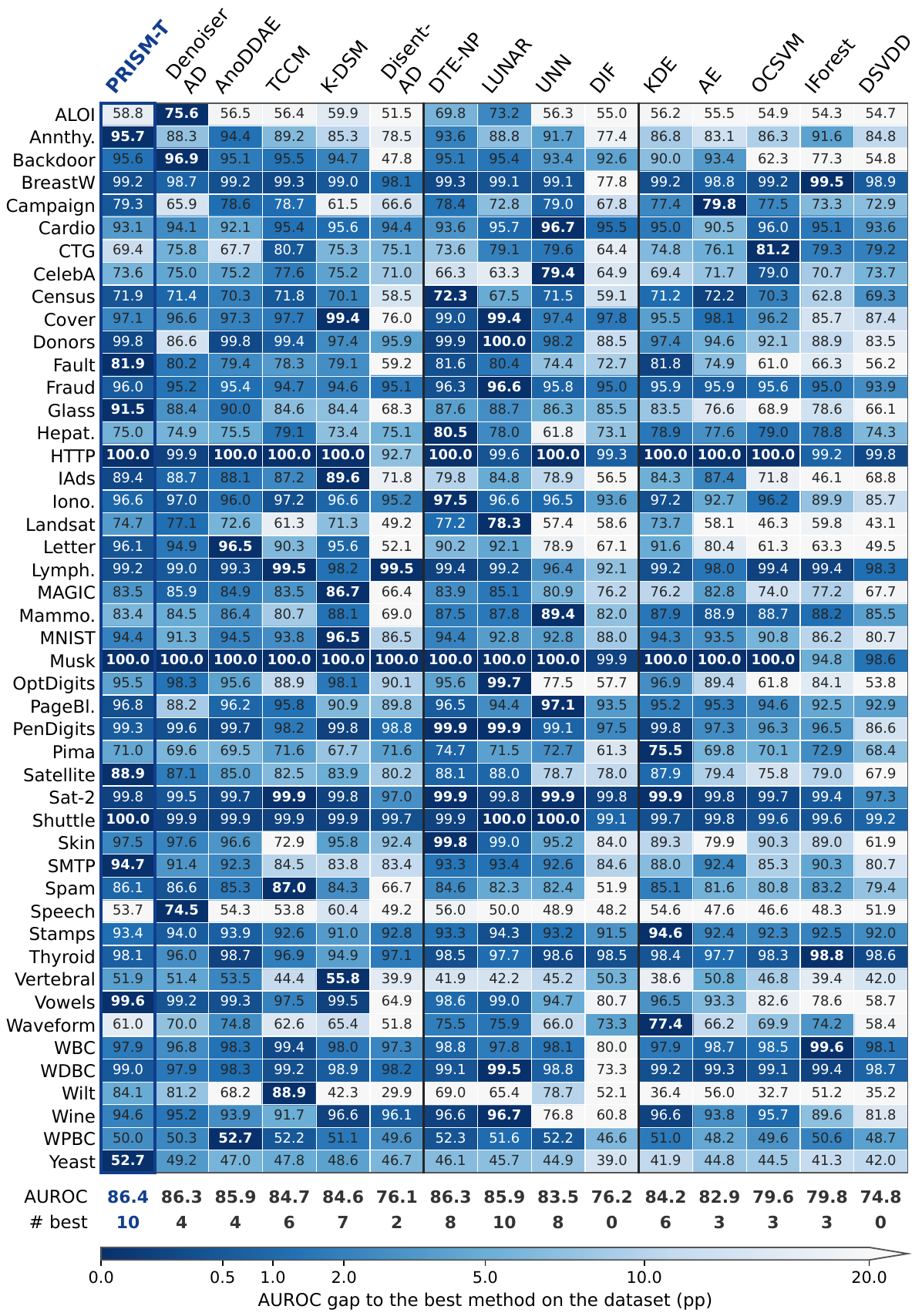}
\caption{
Per-dataset AUROC on ADBench.
Cell values show AUROC~(\%), and colors indicate the gap from the best-performing method.
}
\label{fig:hmap-auroc}
\end{figure}
\begin{figure}[!t]
\centering
\includegraphics[width=\columnwidth]{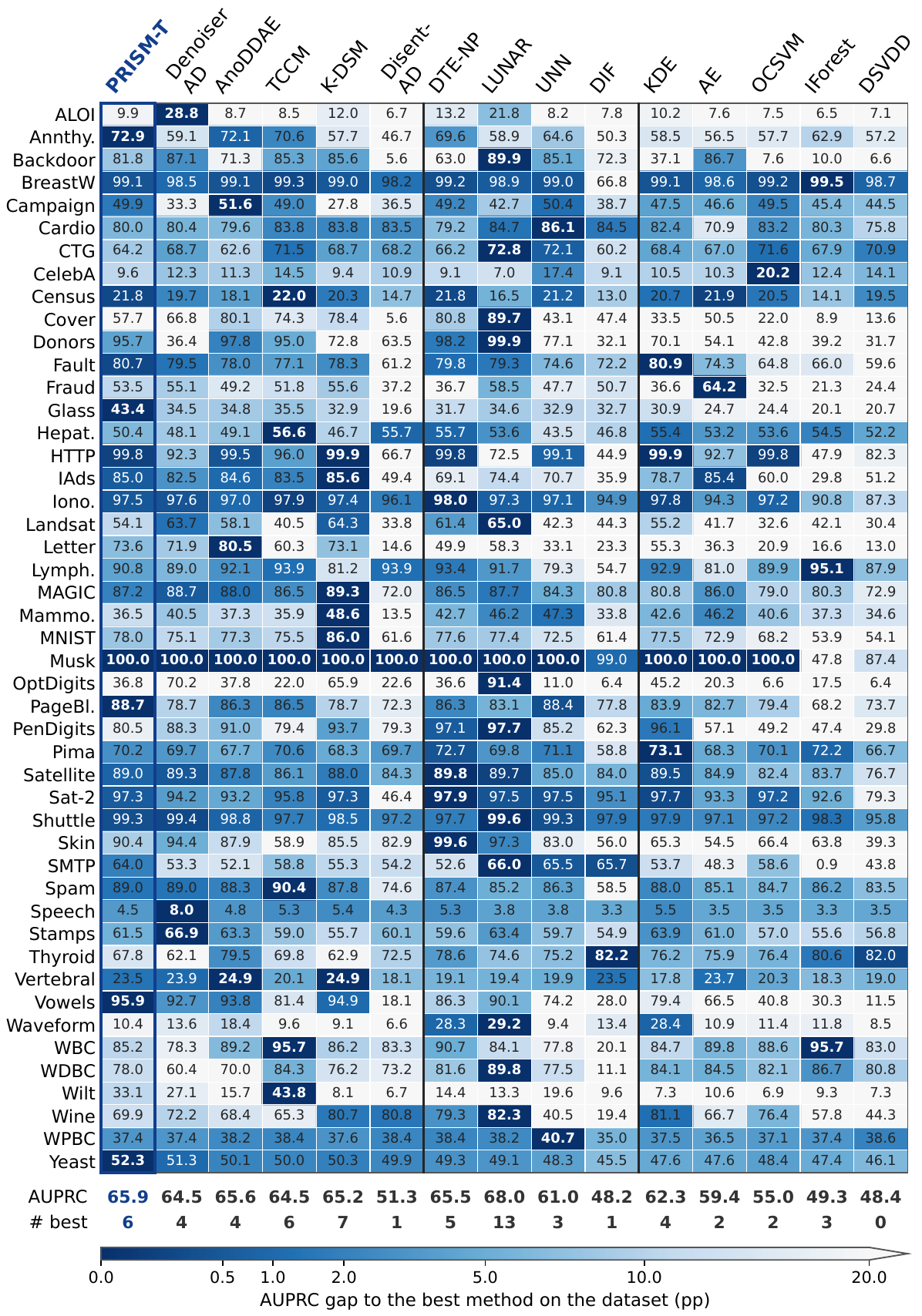}
\caption{Per-dataset AUPRC on ADBench. 
Cell values show AUPRC (\%), and colors indicate the gap from the best-performing method.}
\label{fig:hmap-auprc}
\end{figure}

\begin{figure}[!t]
\centering
\includegraphics[width=\columnwidth]{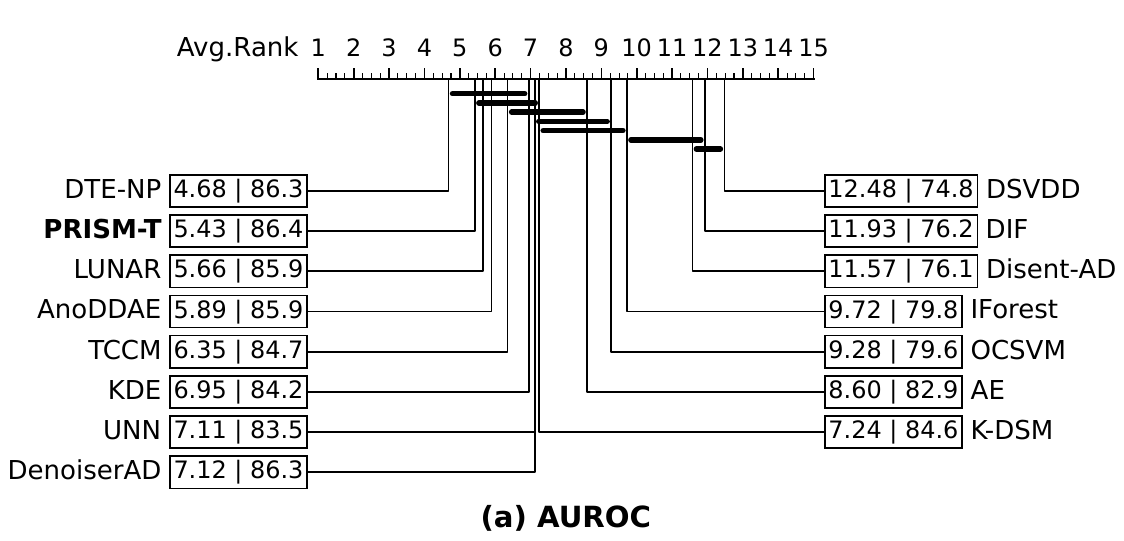}
\includegraphics[width=\columnwidth]{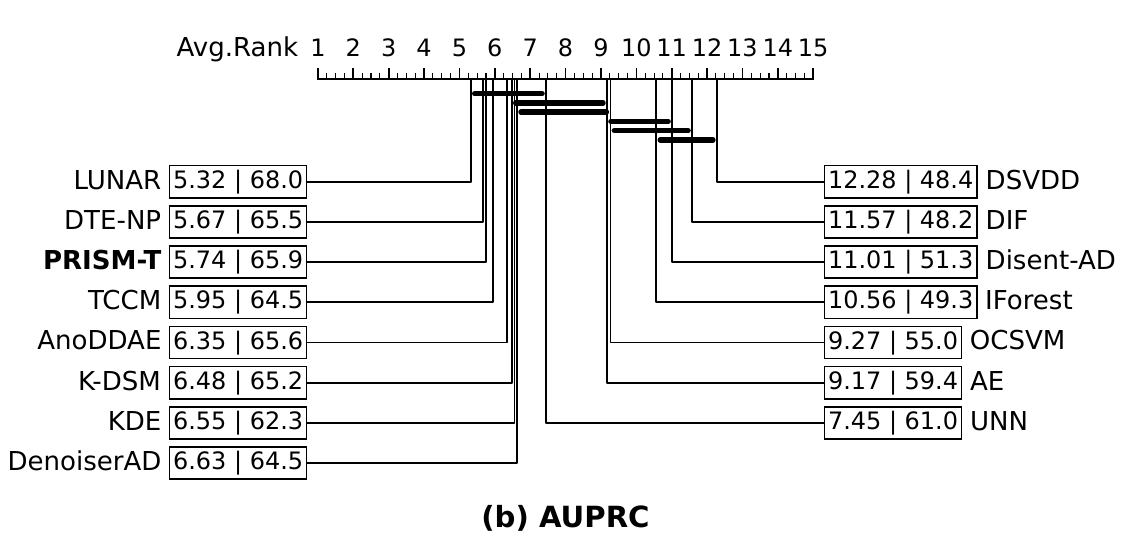}
\caption{Critical-difference diagrams of average ranks on ADBench: (a)~AUROC and (b)~AUPRC.
Numbers indicate average rank and mean score (\%) of each method.
Horizontal bars connect methods with no significant difference (Wilcoxon signed-rank test with Holm correction).
}
\label{fig:cd-tabular}
\end{figure}
\begin{figure}[!t]
    \centering
    \includegraphics[width=\linewidth ]{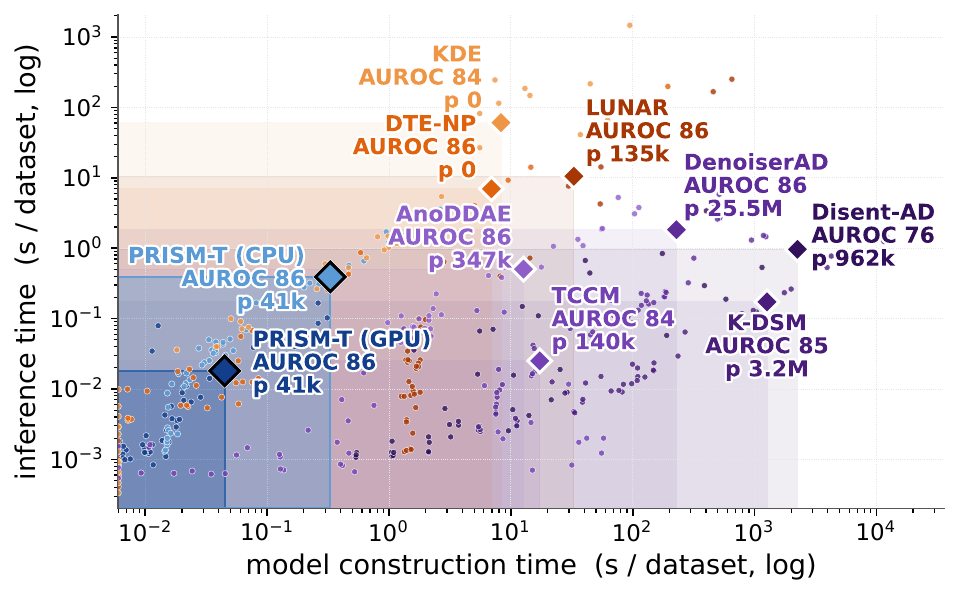}
    \caption{
Construction--inference efficiency on ADBench.
Diamonds denote means and dots denote per-dataset runtimes;
labels show average AUROC and parameter count.
All deep learning baselines ran on GPU, and $p$ denotes the average number of neural network parameters.
}
    \label{fig:efficiency}
\end{figure}

Figs.~\ref{fig:hmap-auroc} and~\ref{fig:hmap-auprc} illustrate per-dataset AUROC and AUPRC performance, respectively.
Although the best-performing method varies across datasets, \textsc{PRISM-T} achieves performance close to the best on many datasets, particularly in AUROC.
This observation is consistent with its leading mean performance and high Top-2 counts reported in Table~\ref{tab:tabular}.

To further examine statistical significance across all methods, Fig.~\ref{fig:cd-tabular} presents critical-difference (CD) diagrams.
The CD diagrams show the average rank of each method, where lower ranks indicate better performance.
Horizontal bars connect groups of methods with no statistically significant pairwise differences under Wilcoxon signed-rank tests with Holm correction over all model pairs ($\alpha=0.05$).
As shown in Fig.~\ref{fig:cd-tabular}, \textsc{PRISM-T} achieves the second-best average rank for AUROC ($5.43$) and the third-best for AUPRC ($5.74$).
Its differences from the top-ranked methods are not statistically significant on either metric.

Beyond detection performance, Fig.~\ref{fig:efficiency} compares the computational efficiency of leading methods in terms of model-construction and inference times. 
Construction time measures the execution of the \texttt{fit()} procedure, while inference time measures the time required to score the test samples.
Existing AD methods exhibit different computational trade-offs.
Nearest-neighbor and kernel-based detectors (e.g., DTE-NP, KDE, and LUNAR) require comparisons with reference samples during inference, leading to increased inference time as the reference set grows.
In contrast, deep learning-based detectors enable efficient inference but require costly weight optimization during model construction. 
LUNAR suffers from both reference-sample comparisons and neural weight optimization, resulting in higher costs in both stages.
\textsc{PRISM-T} avoids both sources of computational overhead by eliminating neural weight optimization and repeated reference-sample comparisons during inference.
As shown in Fig.~\ref{fig:efficiency}, \textsc{PRISM-T} achieves low construction and inference costs.
Its lightweight architecture uses only 41K neural parameters on average, approximately 0.16\%--30.4\% of those in the compared recent neural models.
Based on average execution times across 47 datasets, \textsc{PRISM-T} on CPU achieves $18$--$156\times$ faster inference than the nearest-neighbor and kernel-based detectors, and $38$--$6{,}824\times$ faster model construction than the deep learning detectors.
Moreover, its neural architecture readily supports GPU acceleration through existing deep learning frameworks, further speeding up model construction and inference by $7.4\times$ and $21.6\times$, respectively, compared with its CPU implementation.

Overall, these results demonstrate that \textsc{PRISM-T} combines performance comparable to state-of-the-art methods with low computational costs in both stages, establishing an efficient prior-to-learning baseline for tabular AD.

\begin{table*}[!t]
\caption{Image-level AD and pixel-level anomaly localization on MVTec AD and VisA benchmarks.}
\label{tab:image_localization}
\centering
{
\setlength{\tabcolsep}{5pt}
\begin{tabular}{lc r@{\hspace{1pt}}c r@{\hspace{1pt}}c r@{\hspace{1pt}}c r@{\hspace{1pt}}c c r@{\hspace{1pt}}c r@{\hspace{1pt}}c r@{\hspace{1pt}}c r@{\hspace{1pt}}c c}
\toprule
& & \multicolumn{9}{c}{MVTec AD}
  & \multicolumn{9}{c}{VisA} \\
\cmidrule(lr){3-11}\cmidrule(lr){12-20}
& & \multicolumn{4}{c}{Image-level}
  & \multicolumn{4}{c}{Pixel-level} &
  & \multicolumn{4}{c}{Image-level}
  & \multicolumn{4}{c}{Pixel-level} & \\
\cmidrule(lr){3-6}\cmidrule(lr){7-10}
\cmidrule(lr){12-15}\cmidrule(lr){16-19}
Method & Parameters
& \multicolumn{2}{c}{AUROC}
& \multicolumn{2}{c}{AUPRC}
& \multicolumn{2}{c}{AUROC}
& \multicolumn{2}{c}{AUPRO}
& PRR
& \multicolumn{2}{c}{AUROC}
& \multicolumn{2}{c}{AUPRC}
& \multicolumn{2}{c}{AUROC}
& \multicolumn{2}{c}{AUPRO}
& PRR \\
\midrule

\textbf{PRISM-V}
& 0.33 M
& 92.9 & $\pm 0.2$ & 97.2 & $\pm 0.1$
& 96.2 & $\pm 0.1$ & 85.9 & $\pm 0.2$ & 94.1
& 87.3 & $\pm 0.1$ & 89.1 & $\pm 0.1$
& 96.7 & $\pm 0.0$ & 73.1 & $\pm 0.3$ & 87.8 \\

\midrule

VAE
& 11.1 M
& 69.1 & $\pm 0.6$ & 86.1 & $\pm 0.4$
& 76.8 & $\pm 0.6$ & 54.3 & $\pm 0.5$ & 72.3
& 78.0 & $\pm 2.1$ & 81.2 & $\pm 1.6$
& 84.4 & $\pm 0.4$ & 59.1 & $\pm 0.8$ & 76.8 \\

GANomaly
& 9.06 M
& 57.9 & $\pm 3.3$ & 81.2 & $\pm 1.5$
& \multicolumn{2}{c}{--}
& \multicolumn{2}{c}{--} & 69.7
& 61.7 & $\pm 4.2$ & 69.3 & $\pm 3.3$
& \multicolumn{2}{c}{--}
& \multicolumn{2}{c}{--} & 66.3 \\

\midrule

SPADE$^\dagger$
& 68.9 M
& 79.9 & & 90.4 &
& 96.8 & & 92.4 & & 91.0
& 79.7 & & 82.6 &
& 97.5 & & 85.9 & & 87.7 \\

PaDiM
& 68.9 M
& 96.7 & $\pm 0.1$ & 98.5 & $\pm 0.1$
& 97.5 & $\pm 0.0$ & 92.3 & $\pm 0.1$ & 97.4
& 90.6 & $\pm 0.1$ & 91.5 & $\pm 0.1$
& 97.5 & $\pm 0.0$ & 83.5 & $\pm 0.1$ & 92.2 \\

PatchCore
& 68.9 M
& \underline{98.9} & $\pm 0.2$
& \textbf{99.7} & $\pm 0.1$
& \underline{97.9} & $\pm 0.0$
& 92.3 & $\pm 0.1$ & 98.4
& 94.7 & $\pm 0.2$ & 96.1 & $\pm 0.1$
& 97.9 & $\pm 0.1$ & 87.7 & $\pm 0.4$ & 95.5 \\

WinCLIP+$^\dagger$
& 208 M
& 97.2 & & 98.6 &
& 95.0 & & 86.8 & & 95.5
& 90.5 & & 92.4 &
& 95.3 & & 81.2 & & 91.2 \\

AnomalyDINO$^\dagger$
& 22.1 M
& \textbf{99.2} & & \textbf{99.7} &
& \underline{97.9} & & \underline{93.9} &
& \underline{98.9}
& \underline{97.2} & & \underline{97.6} &
& \underline{98.6} & & \underline{95.0} &
& \underline{98.6} \\

SubspaceAD$^\dagger$
& 1.14 B
& \textbf{99.2} & & \underline{99.6} &
& \textbf{98.2} & & \textbf{95.7} &
& \textbf{99.3}
& \textbf{98.2} & & \textbf{98.3} &
& \textbf{99.1} & & \textbf{96.9} &
& \textbf{99.7} \\

\bottomrule
\end{tabular}
}
\par\smallskip
\parbox{\linewidth}{\scriptsize
For stochastic methods, we report the mean and standard deviation over five seeds; deterministic baselines are marked with $\dagger$ and evaluated once.
\textit{Parameters} denotes the parameter count of the neural network used by each method.
Performance retention ratio (PRR) is the average ratio of each category- and metric-specific score to the corresponding best score.}
\end{table*}

\begin{figure}[!t]
    \centering
    \includegraphics[width=\columnwidth]{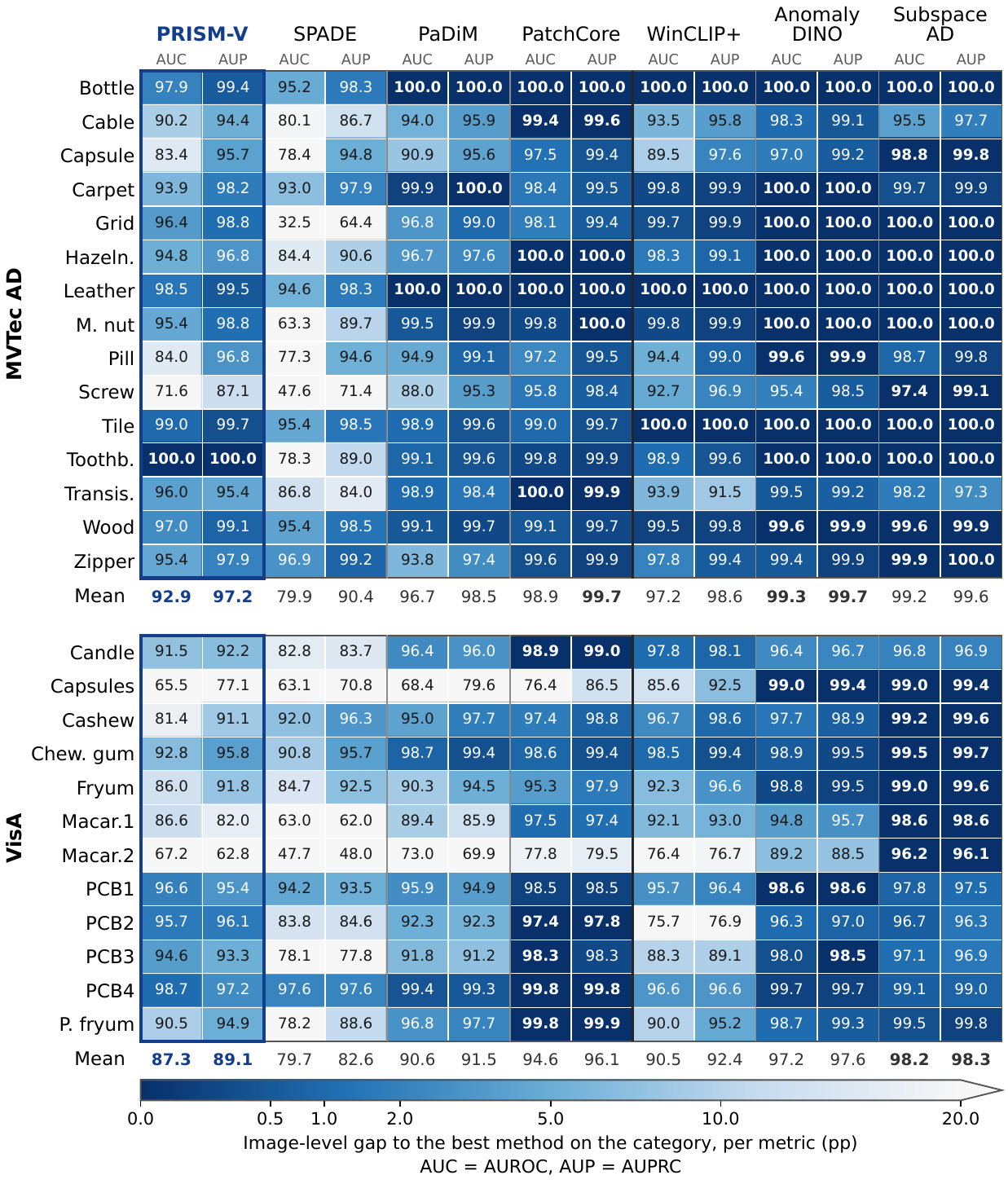}
    \caption{Per-category image-level detection performance on MVTec AD and VisA.
    Cell values show image AUROC (AUC) and AUPRC (AUP) (\%), and colors indicate the gap from the best-performing method.}
    \label{fig:percategory_image}    
\end{figure}

\begin{figure}[!t]
    \centering
    \includegraphics[width=\columnwidth]{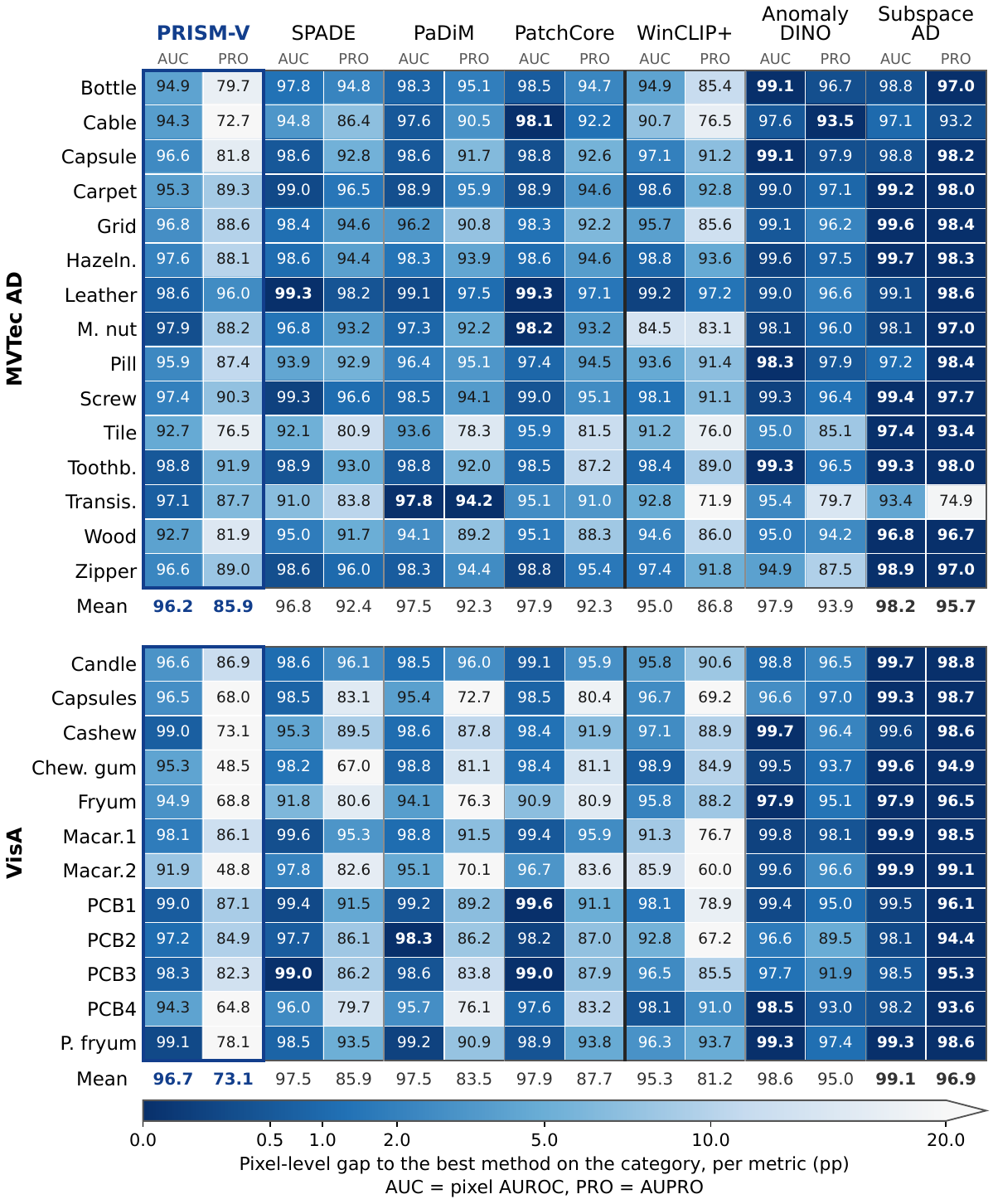}
\caption{Per-category pixel-level localization performance on MVTec AD and VisA.
Cell values show pixel AUROC (AUC) and AUPRO (PRO) (\%), and colors indicate the gap from the best-performing method.}    \label{fig:percategory_pixel}    
\end{figure}

\subsection{Visual Anomaly Detection}
Table~\ref{tab:image_localization} presents image-level detection and pixel-level localization results on MVTec AD and VisA.
Despite relying entirely on randomly initialized networks, \textsc{PRISM-V} achieves image-level AUROC scores of $92.9$ and $87.3$, with corresponding AUPRC scores of $97.2$ and $89.1$, respectively.
This capability extends to dense anomaly localization, achieving pixel-level AUROC scores of $96.2$ and $96.7$ and AUPRO scores of $85.9$ and $73.1$.
\textsc{PRISM-V} substantially outperforms earlier deep learning methods (VAE and GANomaly) on the reported metrics, although a performance gap remains relative to recent pretrained-feature methods.
Among the four evaluation metrics, \textsc{PRISM-V} is most competitive in pixel-level AUROC, showing the smallest gap to the best-performing method on both benchmarks.
To quantify its relative performance across categories and metrics, we use the performance retention ratio (PRR), defined as the average ratio of each category- and metric-specific score to the corresponding best score.
\textsc{PRISM-V} achieves PRRs of $94.1\%$ on MVTec AD and $87.8\%$ on VisA, comparable to those of some pretrained-feature methods (SPADE and WinCLIP+).
Notably, these results are achieved with entirely random neural weights and only 0.33M parameters, less than 0.5\% of those in the pretrained neural networks used by SPADE and WinCLIP+.
This suggests that substantial visual anomaly evidence can be exploited through random nonlinear features and Mahalanobis scoring alone, without learned visual representations.

To examine performance variations across object categories, Figs.~\ref{fig:percategory_image} and~\ref{fig:percategory_pixel} present the per-category results.
At the image level, \textsc{PRISM-V} achieves AUROC scores above $90$ in 12 of the 15 MVTec AD categories, including a perfect score on \textit{Toothbrush}.
It also outperforms PaDiM and WinCLIP+ on \textit{PCB1}--\textit{PCB3} in VisA, although performance varies across categories, particularly on \textit{Screw}, \textit{Capsules}, and \textit{Macaroni2}.
At the pixel level, all 27 categories achieve AUROC scores above $90$, with 20 categories scoring at least $95$.
\textsc{PRISM-V} also achieves higher pixel-level AUROC than multiple pretrained-feature methods in several categories, surpassing five of the six compared methods on \textit{Transistor} and four on \textit{Cashew}. 
At the region level, AUPRO provides a complementary assessment by measuring the coverage of individual anomaly regions.
On \textit{Transistor}, \textsc{PRISM-V} surpasses four pretrained-feature methods in AUPRO.
In contrast, \textit{Cashew} and \textit{Pipe fryum} achieve pixel-level AUROC scores of $99.0$ and $99.1$, respectively, while their corresponding AUPRO scores are $73.1$ and $78.1$.

\begin{figure*}[!ht]
\centering
\includegraphics[width=\textwidth]{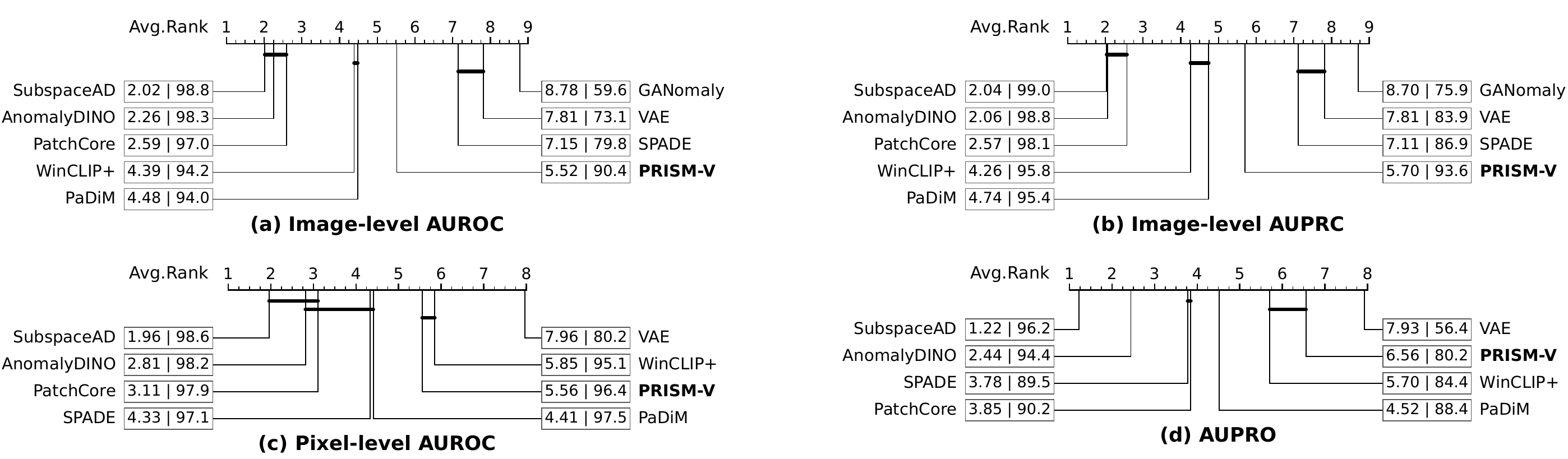}
\caption{
Critical-difference diagrams of average ranks across the 27 categories (MVTec AD and VisA combined) for each metric: (a)~image-level AUROC, (b)~image-level AUPRC, (c)~pixel-level AUROC, and (d)~AUPRO. Numbers indicate average rank and mean score (\%) of each method. 
Horizontal bars connect methods with no significant difference (Wilcoxon signed-rank test with Holm correction).}
\label{fig:cd-visual}
\end{figure*}

Figure~\ref{fig:cd-visual} presents CD diagrams comparing \textsc{PRISM-V} with the baselines using the same statistical procedure as in the tabular AD evaluation.
Although \textsc{PRISM-V} ranks below the leading pretrained-feature methods, it achieves better average ranks than SPADE in image-level detection and WinCLIP+ in pixel-level AUROC.
These findings demonstrate that prior-to-learning anomaly detection extends beyond tabular data to visual detection and pixel-level localization, while also suggesting that pretrained visual representations offer additional advantages.

\section{Further Analysis}
\label{sec6.analysis}

\subsection{Random Transformation and Nonlinearity}

\begin{table}[!t]
\caption{Effect of random transformation and nonlinear activation in \textsc{PRISM}.}
\label{tab:random-activation}
\centering
\setlength{\tabcolsep}{1mm}
\begin{tabular}{@{}lccc cccc@{}}
\toprule
& \multicolumn{3}{c}{Tabular}
& \multicolumn{4}{c}{Visual} \\
\cmidrule(lr){2-4}
\cmidrule(lr){5-8}
Variant
& AUC
& AUP
& $G$
& I-AUC
& P-AUC
& PRO
& $G_{img}/G_{pix}$ \\
\midrule
Raw
& 82.4 & 56.8 & 0.78
& 77.7 & 84.1 & 62.4 & 0.47/0.50 \\
w/o Act.
& 83.2 & 59.3 & 0.77
& 87.9 & 95.0 & 75.4 & 0.58/0.71 \\
\textbf{PRISM}
& \textbf{86.4} & \textbf{65.9} & \textbf{0.80}
& \textbf{90.4} & \textbf{96.4} & \textbf{80.2}
& \textbf{0.74}/\textbf{0.78} \\
\bottomrule
\end{tabular}
\par\smallskip
\parbox{\linewidth}{\scriptsize
``Raw'' uses observation-space Mahalanobis scoring;
``w/o Act.'' removes nonlinear activation layer of PRISM.
}
\end{table}

The preceding results demonstrate the competitive performance of \textsc{PRISM} against existing AD methods.
To investigate how random nonlinear transformations affect anomaly detection performance, Table~\ref{tab:random-activation} compares Mahalanobis scoring in the observation space (Raw), \textsc{PRISM} without nonlinear activation (w/o Act.), and the full model (PRISM).
The Raw variant for visual AD uses pixel-unshuffled RGB inputs to match the scoring grid of \textsc{PRISM-V}, while retaining the same scoring pipeline.

For tabular AD, random transformation improves AUROC and AUPRC over Raw.
Introducing LeakyReLU nonlinearity further improves AUROC by $3.8\%$ and AUPRC by $11.1\%$, with a more pronounced improvement in AUPRC.
The score-level separability $G$ also increases from $0.77$ to $0.80$.
For visual AD, substantial improvements are observed even without nonlinear activation, with image-level AUROC, pixel-level AUROC, and AUPRO increasing by $13.1\%$, $13.0\%$, and $20.8\%$ over Raw, respectively.
LeakyReLU further enhances detection and localization, particularly AUPRO, which rises from $75.4\%$ to $80.2\%$.
This improvement is accompanied by increased score-level separability at both the image and pixel levels, reaching $G_{\mathrm{img}}=0.74$ and $G_{\mathrm{pix}}=0.78$.

Overall, these results demonstrate that random nonlinear representations in \textsc{PRISM} improve anomaly detection over observation-space scoring in both modalities, consistent with the mechanisms discussed in Section~\ref{sec4.theory}. 
The associated changes in score-level separability are examined in the following analyses.

\subsection{Analysis of PRISM-T}
\paragraph{Feature Transformations and Scoring Rules:}
The \textsc{PRISM} framework consists of two components: untrained neural feature transformation and Mahalanobis scoring.
To examine their respective contributions, we compare alternative nonlinear random features and anomaly scoring rules while keeping the other component fixed.

\begin{table}[!t]
\caption{Comparison of feature maps and scoring rules
on ADBench datasets.}
\label{tab:feature_scorer_comparison}
\centering
\begingroup
\setlength{\tabcolsep}{2pt}
\begin{tabular}{lcc@{\hspace{8pt}}lcc}
\toprule
\multicolumn{3}{c}{(a) Feature map}
& \multicolumn{3}{c}{(b) Scoring rule} \\
\cmidrule(lr){1-3}\cmidrule(lr){4-6}
Map & AUROC & AUPRC
& Scorer & AUROC & AUPRC \\
\midrule
Fourier
& $85.9 \pm 0.7$
& $63.8 \pm 2.5$
& Euclidean
& $78.6 \pm 1.5$
& $52.6 \pm 3.0$ \\

Maclaurin
& $84.4 \pm 0.7$
& $61.1 \pm 2.5$
& IForest
& $81.1 \pm 2.3$
& $52.4 \pm 4.8$ \\

ReLU
& $\underline{86.0} \pm 0.7$
& $\underline{64.5} \pm 2.4$
& $k$NN
& $\underline{85.4} \pm 0.9$
& $\underline{63.9} \pm 2.7$ \\

\textbf{PRISM}
& $\mathbf{86.4} \pm 0.9$
& $\mathbf{65.9} \pm 2.5$
& \textbf{Mahal.}
& $\mathbf{86.4} \pm 0.9$
& $\mathbf{65.9} \pm 2.5$ \\
\bottomrule
\end{tabular}%

\endgroup

\par\smallskip
\parbox{\columnwidth}{\scriptsize
(a) Random nonlinear feature transformations are
compared under fixed Mahalanobis scoring.
(b) Anomaly scoring methods are compared using fixed
\textsc{PRISM-T} features.
}
\end{table}

Table~\ref{tab:feature_scorer_comparison}(a) compares \textsc{PRISM-T} with three existing random nonlinear
feature transformations: Fourier~\cite{rahimi2007random}, Maclaurin~\cite{kar2012random}, and ReLU~\cite{cho2009kernel}.
Note that the ReLU feature map differs from simply replacing the activation function in \textsc{PRISM-T}, as it uses a bias-free Gaussian random projection without normalization.
All transformations produce 512-dimensional features and use the same scoring procedure.
Under these conditions, \textsc{PRISM-T} achieves
the highest mean AUROC and AUPRC, followed by ReLU.
Table~\ref{tab:feature_scorer_comparison}(b) compares scoring rules applied to the same PRISM-T features.
Mahalanobis scoring achieves the highest mean AUROC and AUPRC, exceeding $k$NN by 1.0 and 2.0 percentage points (pp), respectively.
While $k$NN achieves comparable performance, it requires nearest-neighbor search over the reference samples at inference.
Mahalanobis scoring instead uses a precomputed mean and precision matrix, making its per-sample scoring cost independent of the reference-set size.
Its substantial advantage over Euclidean scoring further highlights the value of accounting for covariance in the feature space.

Together, the two comparisons support the combination of random neural feature transformation and Mahalanobis scoring in \textsc{PRISM-T}.
The architecture provides an effective nonlinear representation, while covariance-aware scoring translates deviations in that representation into anomaly scores. 

\paragraph{Score-Level Separability:}
This analysis examines how the random neural representation in \textsc{PRISM-T} changes normal--abnormal score separation and how these changes relate to detection performance.
Following Section~\ref{sec4.theory}, the covariance and mean contributions to the score gap are examined together with score dispersion.
Tabular AD provides a direct setting for this analysis, as each sample receives a single anomaly score corresponding to the class-conditional score distributions considered in the theory.

\begin{table}[!t]
\caption{Decomposition of score-level separability on
ADBench datasets.}
\label{tab:score-decomposition}
\centering
\begingroup
\setlength{\tabcolsep}{3pt}
\newcommand{\meanmed}[2]{#1 (#2)}
\begin{tabular}{lccccc}
\toprule
& \multicolumn{3}{c}{Mean (median)}
& \multicolumn{2}{c}{Increase count} \\
\cmidrule(lr){2-4}\cmidrule(lr){5-6}
Quantity & Raw & w/o Act. & PRISM
& \shortstack{vs.\\Raw}
& \shortstack{vs.\\w/o Act.} \\
\midrule
$G$
& \meanmed{0.78}{0.39}
& \meanmed{0.77}{0.36}
& \meanmed{0.80}{0.45}
& 34 & 35 \\

$G^{\mathrm{cov}}$
& \meanmed{0.24}{0.18}
& \meanmed{0.28}{0.20}
& \meanmed{0.49}{0.34}
& 40 & 39 \\

$G^{\mathrm{mean}}$
& \meanmed{0.55}{0.15}
& \meanmed{0.49}{0.09}
& \meanmed{0.31}{0.11}
& 19 & 25 \\
\midrule
$\Delta$
& \meanmed{46.88}{0.06}
& \meanmed{114.82}{0.42}
& \meanmed{91.50}{4.11}
& 41 & 36 \\

$\Delta^{\mathrm{cov}}$
& \meanmed{45.18}{0.04}
& \meanmed{110.56}{0.21}
& \meanmed{60.58}{3.12}
& 38 & 35 \\

$\Delta^{\mathrm{mean}}$
& \meanmed{1.69}{0.02}
& \meanmed{4.17}{0.07}
& \meanmed{29.33}{0.75}
& 44 & 39 \\

$\sqrt{V_r}$
& \meanmed{203.15}{0.13}
& \meanmed{679.81}{0.64}
& \meanmed{256.58}{12.45}
& 41 & 39 \\
\bottomrule
\end{tabular}
\endgroup

\par\smallskip
\parbox{\columnwidth}{\scriptsize
Entries report means, with medians in parentheses.
Increase counts indicate the number of datasets with higher values than the baseline (out of 47).
Gap and dispersion values ($\Delta$ and $\sqrt{V_r}$)
are reported in units of $10^3$.
}
\end{table}

\begin{figure}[!t]
    \centering
    \includegraphics[width=\columnwidth]{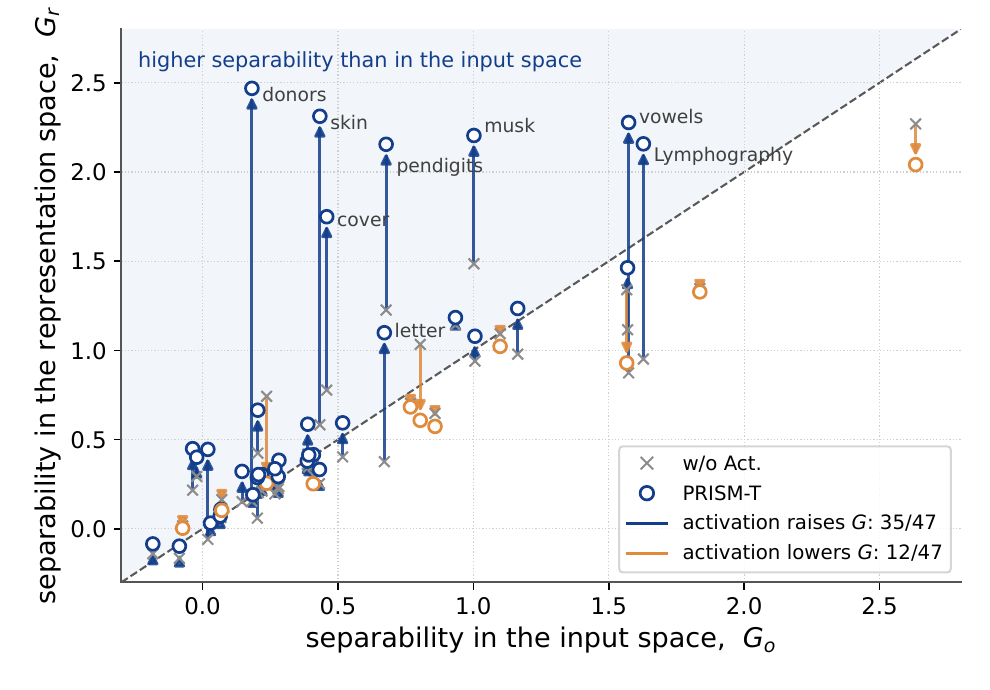}
\caption{Score-level separability across 47 ADBench datasets.
Crosses and circles indicate w/o Act. and \textsc{PRISM-T}, respectively; arrows show changes induced by nonlinear activation.
The dashed line denotes $G_r=G_o$.} 
    \label{fig:G-separability}
\end{figure}

Using anomaly scores from normal and anomalous test samples, Table~\ref{tab:score-decomposition} summarizes $G$, $\Delta$, and $\sqrt{V}$, including the mean and covariance contributions to $\Delta$, for Raw, w/o Act., and \textsc{PRISM-T}.
The unstandardized mean contribution $\Delta^{\mathrm{mean}}$ increases on 44 and 39 of the 47 datasets relative to Raw and w/o Act., respectively, with increases in both its mean and median.
This pattern is consistent with the theoretical role of nonlinear transformation in enlarging the mean contribution to the score gap.
After normalization by score dispersion, however, the covariance contribution shows more consistent gains: $G^{\mathrm{cov}}$ increases on 40 and 39 datasets relative to Raw and w/o Act., whereas $G^{\mathrm{mean}}$ increases on 19 and 25 datasets.
The resulting changes in overall score-level separability $G$ are illustrated across individual datasets in Fig.~\ref{fig:G-separability}.
\textsc{PRISM-T} improves separation across most datasets, although the magnitude of improvement varies.

\begin{figure}[!t]
\centering
\includegraphics[width=\columnwidth]{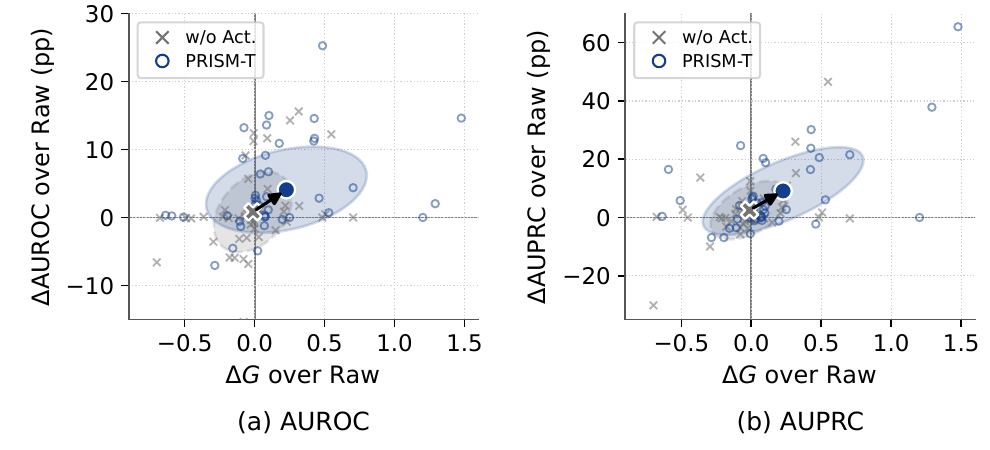}
\caption{
Changes in separability and detection performance relative to Raw
on 47 datasets: (a) AUROC and (b) AUPRC.
Large markers and ellipses indicate the mean and $\pm$ one standard deviation across datasets, excluding \textit{http} due to its outlying $\Delta G$.
}
\label{fig:g-performance}
\end{figure}

Fig.~\ref{fig:g-performance} relates changes in score-level separability $G$ to changes in AUROC and AUPRC relative to Raw.
Without nonlinear activation, the random transformation produces little change in average separability ($\Delta G=-0.03$), with modest improvements in AUROC (+0.9\,pp) and AUPRC (+2.6\,pp).
In contrast, adding LeakyReLU shifts the average toward higher separability ($\Delta G=+0.23$) and larger improvements in AUROC (+4.1\,pp) and AUPRC (+9.1\,pp).
Across datasets, changes in $G$ from Raw to \textsc{PRISM-T} are positively correlated with the corresponding changes in AUROC and AUPRC (Spearman's $\rho=0.49$ and $0.47$, respectively).

The decomposition and performance analyses show that random nonlinear transformations modify both the score gap and its dispersion, with distinct effects on the mean and covariance contributions.
The resulting improvements in score-level separability are associated with improved anomaly detection performance, supporting the role of random nonlinear representations in exposing intrinsic anomaly separability.

\subsection{Analysis of PRISM-V}

\begin{table}[!t]
\caption{
Comparison with randomly initialized baselines.
}
\label{tab:random_init_baselines}
\centering
\begingroup
\resizebox{\columnwidth}{!}{%
\begin{tabular}{llcc}
\toprule
Method & Backbone & I-AUROC & AUPRO \\
\midrule
\textbf{PRISM-V}
& Proposed Backbone
& $\mathbf{90.4} \pm 0.1$
& $\mathbf{80.2} \pm 0.1$ \\
\midrule
SPADE
& \multirow{3}{*}{WRN-50-2}
& $60.8 \pm 0.7$
& $52.0 \pm 1.0$ \\

PaDiM
&
& $\underline{79.5} \pm 0.1$
& $55.5 \pm 0.5$ \\

PatchCore
&
& $76.6 \pm 1.5$
& $51.3 \pm 1.2$ \\
\midrule
WinCLIP+
& \begin{tabular}[c]{@{}l@{}}
V: CLIP ViT-B/16+\\
L: CLIP Transformer
\end{tabular}
& $73.0 \pm 0.9$
& $43.5 \pm 4.5$ \\
\midrule
AnomalyDINO
& DINOv2 ViT-S/14
& $77.3 \pm 0.4$
& $\underline{60.2} \pm 0.6$ \\

SubspaceAD
& DINOv2 ViT-g/14
& $68.5 \pm 0.4$
& $54.4 \pm 0.4$ \\
\bottomrule
\end{tabular}
}

\par\smallskip
\parbox{\columnwidth}{\scriptsize
WRN-50-2 denotes Wide ResNet-50-2.
For WinCLIP+, V and L denote the vision and language encoders.
ViT-S/14 and ViT-g/14 denote the small and giant Vision Transformers, respectively.
}
\endgroup
\end{table}

\paragraph{Feature Transformations and Scoring Rules:}
To further investigate the effectiveness of random feature representations in visual AD, we replace the pretrained backbone originally used by each baseline with a randomly initialized counterpart, while retaining its anomaly scoring procedure.
These backbones include Wide ResNet~\cite{zagoruyko2016wide}, CLIP~\cite{radford2021clip}, and DINOv2~\cite{oquab2024dinov2}, as detailed in Table~\ref{tab:random_init_baselines}.
Across the six randomly initialized baselines, image-level AUROC ranges from 60.8 to 79.5, while AUPRO ranges from 43.5 to 60.2.
In contrast, \textsc{PRISM-V} achieves 90.4 and 80.2, exceeding the strongest baselines by 10.9\,pp and 20.0\,pp, respectively.
Notably, PaDiM achieves the highest image-level AUROC among the baselines and employs a similar scoring principle based on location-specific Mahalanobis distance, yet remains substantially below \textsc{PRISM-V}.
These results support the effectiveness of \textsc{PRISM-V}'s architectural design in exploiting random neural features for visual anomaly detection.

Table~\ref{tab:visual_scorer_comparison} compares four scoring rules (Euclidean, IForest, $k$NN, and Mahalanobis) applied to the same fixed 512-dimensional \textsc{PRISM-V} features.
All scorers are fitted location-wise.
Consistent with \textsc{PRISM-T}, Mahalanobis scoring achieves the highest mean performance across all four metrics, followed by $k$NN. 
The advantage is more pronounced in \textsc{PRISM-V}, where Mahalanobis outperforms $k$NN by 2.2--9.2\,pp across the four metrics.
Compared with Euclidean scoring, it improves image-level AUROC and AUPRO by 19.2\,pp and 20.3\,pp, respectively.
The larger performance gains in \textsc{PRISM-V} indicate a greater benefit from covariance-aware scoring when using random visual features.

\begin{table}[!t]
\caption{
Comparison of scoring rules for PRISM-V.
}
\label{tab:visual_scorer_comparison}
\centering
\begingroup
\setlength{\tabcolsep}{4pt}
\begin{tabular}{lcccc}
\toprule
Scorer & I-AUROC & I-AUPRC & P-AUROC & AUPRO \\
\midrule

Euclidean
& $71.2 \pm 0.3$
& $81.0 \pm 0.3$
& $89.0 \pm 0.1$
& $59.9 \pm 0.4$ \\

IForest
& $79.3 \pm 0.2$
& $85.9 \pm 0.1$
& $89.4 \pm 0.2$
& $58.0 \pm 0.2$ \\

$k$NN
& $\underline{81.2} \pm 0.3$
& $\underline{88.0} \pm 0.2$
& $\underline{94.2} \pm 0.1$
& $\underline{74.5} \pm 0.2$ \\

\textbf{Mahalanobis}
& $\mathbf{90.4} \pm 0.1$
& $\mathbf{93.6} \pm 0.0$
& $\mathbf{96.4} \pm 0.0$
& $\mathbf{80.2} \pm 0.1$ \\
\bottomrule
\end{tabular}

\endgroup

\end{table}

\begin{table}[!t]
\caption{Score-level separability of PRISM-V on
MVTec AD and VisA.}
\label{tab:score-separability-visual}
\centering
\begingroup
\setlength{\tabcolsep}{4pt}
\newcommand{\meanmed}[2]{#1 (#2)}
\begin{tabular}{lccccc}
\toprule
& \multicolumn{3}{c}{Mean (median)}
& \multicolumn{2}{c}{Increase count} \\
\cmidrule(lr){2-4}\cmidrule(lr){5-6}
Quantity & Raw & w/o Act. & PRISM-V
& \shortstack{vs.\\Raw}
& \shortstack{vs.\\w/o Act.} \\
\midrule
$G_{\mathrm{img}}$
& \meanmed{0.47}{0.44}
& \meanmed{0.58}{0.52}
& \meanmed{0.74}{0.63}
& 22 & 22 \\

$G_{\mathrm{pix}}$
& \meanmed{0.50}{0.48}
& \meanmed{0.71}{0.69}
& \meanmed{0.78}{0.77}
& 22 & 16 \\
\midrule
$\Delta_{\mathrm{img}}$
& \meanmed{0.05}{0.02}
& \meanmed{2.26}{1.44}
& \meanmed{6.63}{3.51}
& 27 & 26 \\

$\Delta_{\mathrm{pix}}$
& \meanmed{0.04}{0.01}
& \meanmed{1.79}{1.13}
& \meanmed{4.68}{2.40}
& 27 & 25 \\
\midrule
$\sqrt{V_{\mathrm{img}}}$
& \meanmed{0.14}{0.06}
& \meanmed{3.70}{2.56}
& \meanmed{8.23}{5.51}
& 27 & 24 \\

$\sqrt{V_{\mathrm{pix}}}$
& \meanmed{0.09}{0.04}
& \meanmed{2.73}{1.65}
& \meanmed{6.17}{3.76}
& 27 & 25 \\
\bottomrule
\end{tabular}
\endgroup

\par\smallskip
\parbox{\columnwidth}{\scriptsize
Entries report means, with medians in parentheses.
Increase counts indicate the number of categories with higher values than the baseline (out of 27).}
\end{table}

\begin{figure}[!t]
    \centering
    \includegraphics[width=\columnwidth]{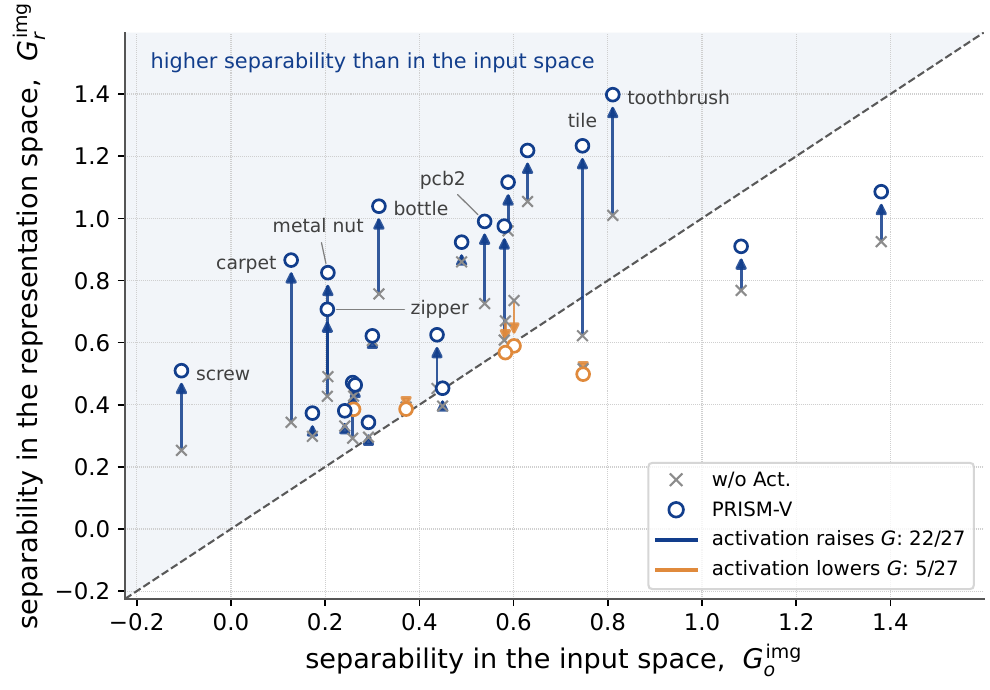}
    \par
    {\small (a) Image-level separability per category\par}
    
    \vspace{1mm}

    \includegraphics[width=\columnwidth]{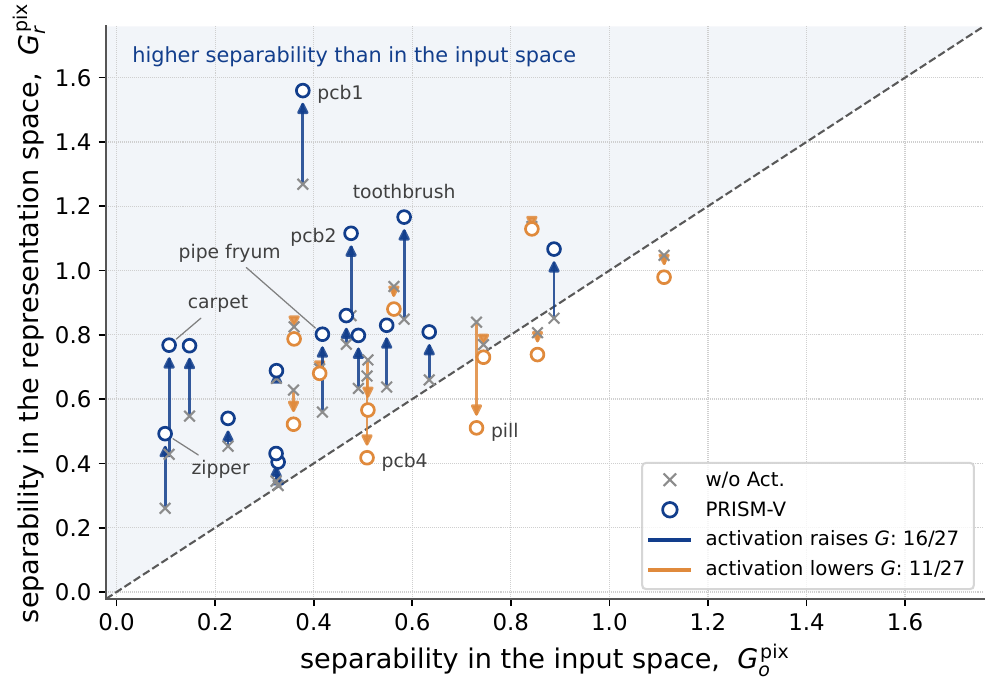}
    \par
    {\small (b) Pixel-level separability per category\par}

\caption{Score-level separability across 27 categories of MVTec AD and VisA.
Crosses and circles indicate w/o Act. and \textsc{PRISM-V}, respectively; arrows show changes induced by nonlinear activation.
The dashed line denotes $G_r=G_o$.}
    \label{fig:G-separability-v}
\end{figure}

\begin{figure}[!t]
\centering
\includegraphics[width=\columnwidth]{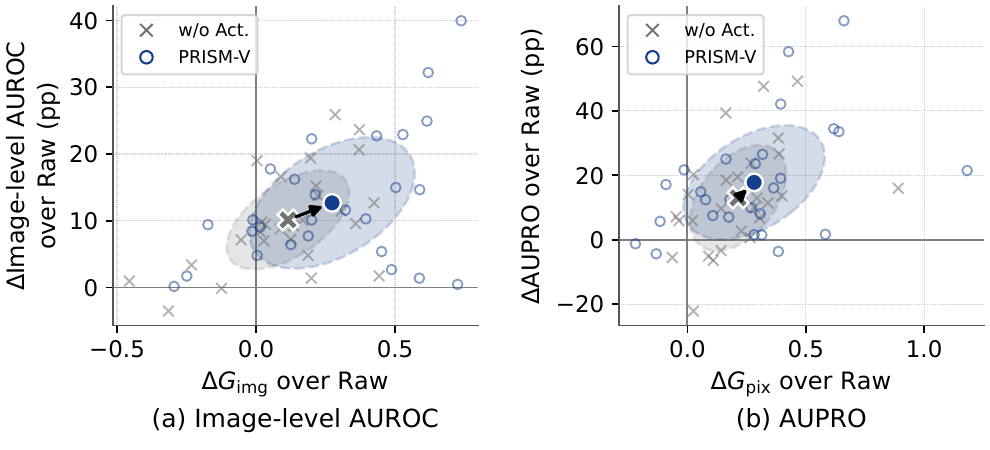}
\caption{
Changes in separability and detection performance relative to Raw
on the 27 categories of MVTec AD and VisA:
(a) image-level AUROC versus $\Delta G_{\mathrm{img}}$ and
(b) AUPRO versus $\Delta G_{\mathrm{pix}}$.
Large markers and ellipses indicate the mean and $\pm$ one standard deviation over the 27 categories, and the arrow connects the two means.
}
\label{fig:g-performance-v}
\end{figure}

\paragraph{Score-Level Separability:}
We extend the score-level separability analysis to visual AD across the 27 categories of MVTec AD and VisA.
While the theoretical formulation defines $G$ using squared Mahalanobis distances, visual AD involves spatial postprocessing and aggregation of Mahalanobis distances.
Accordingly, we evaluate separability using the squared pixel-level anomaly scores $S_{\mathrm{pix}}(I)=(A_{\mathrm{pix}}(I))^2$ and its image-level aggregation $S_{\mathrm{img}}(I)=
\operatorname{TVaR}_{1\%}(S_{\mathrm{pix}}(I))$.
Normal and abnormal score distributions are constructed by partitioning pixel-level scores using ground-truth masks and image-level scores using image labels.
Since the resulting scores no longer retain the exact quadratic form of the original Mahalanobis score, we analyze $G$, $\Delta$, and $\sqrt{V}$ without decomposing the score gap into mean and covariance contributions.

Table~\ref{tab:score-separability-visual} presents the image- and pixel-level separability results for Raw, w/o Act., and \textsc{PRISM-V}.
As in tabular AD, \textsc{PRISM-V} improves score-level separability, but the gains are more pronounced, with $G$ increasing at both stages of feature transformation.
Mean $G_{\mathrm{img}}$ increases from 0.47 for Raw and 0.58 for w/o Act. to 0.74, with improvements in 22 of 27 categories relative to each baseline.
At the pixel level, mean $G_{\mathrm{pix}}$ increases from 0.50 and 0.71 to 0.78.
A notable change is observed in the score gap $\Delta$. 
Its mean increases substantially from near-zero values in Raw (0.05 and 0.04) to 6.63 and 4.68 in \textsc{PRISM-V} at the image and pixel levels, respectively.
These correspond to approximately 133- and 117-fold increases, while the mean score dispersion $\sqrt{V}$ increases by 59- and 69-fold.
The category-wise changes in $G$ are further illustrated in Fig.~\ref{fig:G-separability-v}.
Fig.~\ref{fig:g-performance-v} shows changes in $G$ versus changes in detection and localization performance across categories.
Consistent with the analysis of \textsc{PRISM-T}, positive correlations are observed (Spearman's $\rho=0.41$, $0.56$, and $0.51$ for image-level AUROC, pixel-level AUROC, and AUPRO, respectively).

\begin{figure*}[!t]
    \centering
    \includegraphics[width=\textwidth]{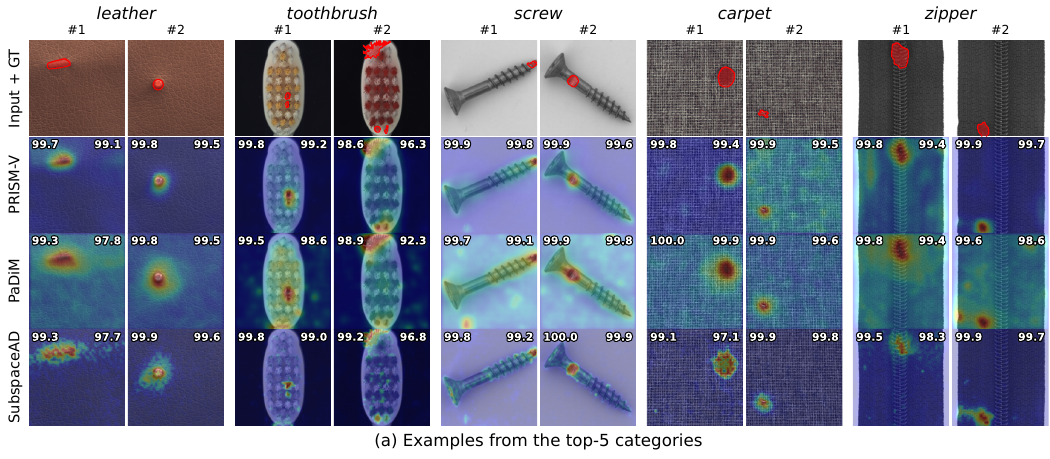}
    \vspace{3pt}
    \includegraphics[width=\textwidth]{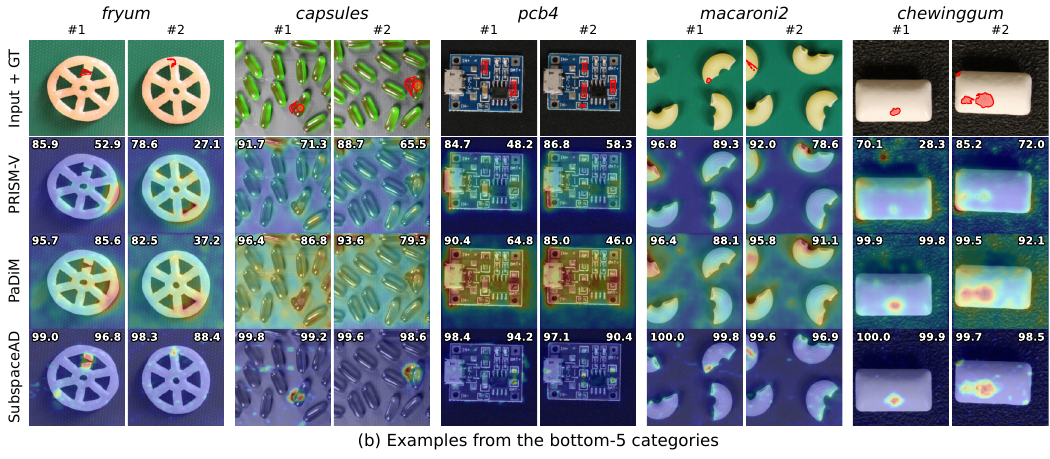}
    \caption{
    Illustrative comparison of anomaly localization using \textsc{PRISM-V}, PaDiM, and SubspaceAD. 
    Examples from the five highest- and lowest-performing categories ranked by \textsc{PRISM-V}'s AUPRO.
    GT denotes ground-truth anomaly regions.
    For each sample, values in the upper-left and upper-right corners indicate pixel AUROC and AUPRO (\%), respectively.    
    }
   
    \label{fig:anomaly_map_sc}
\end{figure*}

\paragraph{Visual Inspection of Anomaly Localization:}
Figure~\ref{fig:anomaly_map_sc} compares the anomaly localization results of \textsc{PRISM-V}, PaDiM, and SubspaceAD across the five highest- and five lowest-AUPRO categories for \textsc{PRISM-V}.
In the top-5 categories, \textsc{PRISM-V} produces concentrated anomaly responses around the ground-truth defects, as illustrated by \textit{leather}, \textit{screw}, and \textit{zipper}.
Both pixel AUROC and AUPRO remain high, with localization performance generally comparable to that of the pretrained-feature baselines.
These results demonstrate that randomly initialized CNN features can support precise anomaly localization without relying on learned visual representations.
In contrast, the bottom-5 categories exhibit less selective anomaly responses, often extending into normal object boundaries or surrounding structures, as observed in \textit{fryum}, \textit{pcb4}, and \textit{chewinggum}.
In categories with multiple similar objects, such as \textit{capsules} and \textit{macaroni2}, anomaly responses also spread beyond the ground-truth defects.
SubspaceAD generally provides more focused localization in these cases.
Even in these challenging categories, \textsc{PRISM-V} exhibits elevated anomaly responses near some ground-truth defects, although these responses are less spatially selective and extend into normal regions.

These visual results complement the quantitative findings by showing that \textsc{PRISM-V} produces localized anomaly responses in favorable categories while retaining sensitivity to defects in more challenging cases.
Together, they further demonstrate the potential of untrained neural features for anomaly localization.

\subsection{Ablation Studies}
\label{subsec:ablation}
While the preceding analyses focused on score-level separability, this section examines how anomaly detection performance varies with model components and hyperparameter settings.
Architectural configurations used in the main experiments are marked with~${*}$.
\begin{table}[!t]
\caption{Modality-specific component ablations for \textsc{PRISM}.}
\label{tab:modality-components}
\centering
{
\setlength{\tabcolsep}{5pt}
\newcommand{\sd}[2]{$#1_{(#2)}$}
\begin{tabular}{llllll}
\toprule
\multicolumn{3}{c}{Tabular}
&
\multicolumn{3}{c}{Visual}
\\
\cmidrule(lr){1-3}
\cmidrule(lr){4-6}
& AUROC & AUPRC
& & I-AUROC & AUPRO \\
\midrule
Base & \sd{86.2}{-} & \sd{64.4}{-} & Base & \sd{82.0}{-} & \sd{69.5}{-} \\
+IN & \sd{80.5}{-5.7} & \sd{54.4}{-10} & +IN & \sd{83.3}{+1.3} & \sd{72.4}{+2.9} \\
+BN & \sd{86.1}{-0.1} & \sd{64.2}{-0.2} & +BP & \sd{83.8}{+1.8} & \sd{75.8}{+6.3} \\
\textbf{+IBN} & \sd{\textbf{86.4}}{+0.2} & \sd{\textbf{65.9}}{+1.5} & \textbf{+Full} & \sd{\textbf{90.4}}{+8.4} & \sd{\textbf{80.2}}{+10.7} \\
\bottomrule
\end{tabular}
\par\smallskip
\parbox{\linewidth}{\scriptsize
IN/BN/IBN denote instance/batch/instance-batch normalization; BP denotes BlurPool.
Visual Base omits normalization and uses MaxPool in place of BP.
Parentheses show differences from Base.}}
\end{table}
Table~\ref{tab:modality-components} summarizes the modality-specific effects of normalization and pooling.
For \textsc{PRISM-T}, IN alone degrades performance, whereas IBN achieves the best results.
For \textsc{PRISM-V}, BlurPool and IN each improve performance, with their combination yielding the largest gains.
Table~\ref{tab:arch-ablation} examines the effects of feature extractor width, depth, and activation functions.
Increasing feature width improves performance in both variants, although the gains tend to diminish at larger widths, while covariance estimation and inversion become more costly.
For \textsc{PRISM-T}, increasing depth provides no consistent improvement, suggesting limited benefits from stacking additional randomly initialized PBs. 
Regarding activation functions, ReLU and LeakyReLU yield nearly identical results, whereas Tanh and Sigmoid generally degrade performance, particularly in visual anomaly localization.
In addition, spatially averaging location-wise covariance estimates over a $3\times3$ neighborhood improves performance in \textsc{PRISM-V}.
Compared with no spatial covariance averaging, image-level AUROC increases by 1.8\,pp and 1.3\,pp on MVTec AD and VisA, respectively.
The corresponding AUPRO gains are 0.2\,pp and 0.5\,pp.

\begin{table}[!t]
\caption{Ablations of feature extractor configurations in \textsc{PRISM}.}
\label{tab:arch-ablation}
\centering
\setlength{\tabcolsep}{3pt}
\begin{tabular}{llcccc}
\toprule
\multicolumn{6}{c}{\textbf{PRISM-T}} \\
\midrule
\textbf{Factor} & \textbf{Configuration}
& \multicolumn{2}{c}{\textbf{AUROC}}
& \multicolumn{2}{c}{\textbf{AUPRC}} \\
\midrule
\multirow{4}{*}{Width}
  & $128$     & \multicolumn{2}{c}{85.61} & \multicolumn{2}{c}{62.83} \\
  & $256$     & \multicolumn{2}{c}{86.01} & \multicolumn{2}{c}{64.83} \\
  & $512^{*}$ & \multicolumn{2}{c}{86.40} & \multicolumn{2}{c}{65.90} \\
  & $1024$    & \multicolumn{2}{c}{86.60} & \multicolumn{2}{c}{66.64} \\
\midrule
\multirow{3}{*}{Depth}
  & $1^{*}$ & \multicolumn{2}{c}{86.40} & \multicolumn{2}{c}{65.90} \\
  & $2$     & \multicolumn{2}{c}{86.21} & \multicolumn{2}{c}{66.33} \\
  & $3$     & \multicolumn{2}{c}{85.93} & \multicolumn{2}{c}{65.74} \\
\midrule
\multirow{4}{*}{Activation}
  & LeakyReLU$^{*}$ & \multicolumn{2}{c}{86.40} & \multicolumn{2}{c}{65.90} \\
  & ReLU            & \multicolumn{2}{c}{86.40} & \multicolumn{2}{c}{65.90} \\
  & Tanh            & \multicolumn{2}{c}{86.30} & \multicolumn{2}{c}{64.63} \\
  & Sigmoid         & \multicolumn{2}{c}{85.68} & \multicolumn{2}{c}{62.66} \\
\midrule
\multicolumn{6}{c}{\textbf{PRISM-V}} \\
\midrule
\textbf{Factor} & \textbf{Configuration}
& \textbf{I-AUC} & \textbf{I-AUP}
& \textbf{P-AUC} & \textbf{PRO} \\
\midrule
\multirow{3}{*}{Width}
  & $32$      & 87.77 & 91.95 & 95.32 & 76.60 \\
  & $64$      & 89.26 & 92.89 & 96.08 & 79.11 \\
  & $128^{*}$ & 90.40 & 93.60 & 96.44 & 80.24 \\
\midrule
\multirow{4}{*}{Activation}
  & LeakyReLU$^{*}$ & 90.40 & 93.60 & 96.44 & 80.24 \\
  & ReLU            & 90.40 & 93.60 & 96.43 & 80.25 \\
  & Tanh            & 88.67 & 92.61 & 94.68 & 73.30 \\
  & Sigmoid         & 86.74 & 91.13 & 94.86 & 74.52 \\
\bottomrule
\end{tabular}
\par\smallskip
\parbox{\linewidth}{\scriptsize
${*}$ denotes the configuration used in the main experiments.}
\end{table}

\section{Conclusion}
\label{sec7.conclusion}
In this work, we introduced \textsc{PRISM}, a prior-to-learning anomaly detection framework that combines fixed random nonlinear features with normal-reference Mahalanobis scoring.
Its core component is a PRISM block that follows the typical linear transformation--normalization--activation pattern in neural networks, using a randomly initialized linear operator and a kinked nonlinearity.
This design is instantiated in two modality-specific detectors: \textsc{PRISM-T} uses a single fully connected PRISM block with mixed instance and batch normalization for tabular anomaly scoring, while \textsc{PRISM-V} employs multi-scale convolutional features and location-wise scoring for visual anomaly localization.

Despite requiring neither pretraining nor neural weight optimization, both models achieve competitive detection performance.
Across 47 tabular datasets in ADBench, \textsc{PRISM-T} achieves the highest mean AUROC (86.4\%) and the second-highest mean AUPRC (65.9\%) among the compared methods, while ranking first on the largest number of datasets in terms of AUROC.
It also avoids stage-specific computational overhead, achieving up to 156$\times$ faster inference than nearest-neighbor and kernel-based detectors and up to 6,824$\times$ faster model construction than recent deep learning detectors.
For visual AD, \textsc{PRISM-V} demonstrates effective detection and localization on MVTec AD and VisA, achieving mean image-level and pixel-level AUROCs of 90.4\% and 96.4\%, respectively.
Across the evaluated metrics, it retains an average of 94.1\% and 87.8\% of the best category- and metric-specific scores on the respective benchmarks, using a fully untrained network with only 0.33M parameters, approximately 0.03\%--1.5\% of those used by the compared pretrained-feature detectors.
Our theoretical analysis further characterizes how nonlinear transformations affect the covariance and mean contributions to the expected Mahalanobis score gap.
Complementary empirical analyses examine the resulting score separability and its relationship with detection performance, while ablation studies assess the effects of key architectural choices.

Together, these findings establish \textsc{PRISM} as a strong and computationally efficient prior-to-learning model across tabular and visual anomaly detection, providing a reference for assessing the added value of learned representations.

\section*{Acknowledgement}
This work was supported in part by the National Research Foundation of Korea (NRF) grant funded by the Korea government (MSIT) (RS-2024-00454514, 50\%), the Institute of Information \& Communications Technology Planning \& Evaluation (IITP) - ICAN (ICT Challenge and Advanced Network of HRD) grant funded by the Korea government (Ministry of Science and ICT) (IITP-2026-RS-2024-00436528, 40\%), and the Institute Strategic Development Project (ISD) in Korea Atomic Energy Research Institute (KAERI) funded by the Ministry of Science and ICT (No. 526140-26, 10\%).

\bibliographystyle{IEEEtran}
\bibliography{references}

\end{document}